\documentclass[10pt,journal,letterpaper,compsoc]{IEEEtran}
\usepackage[nocompress]{cite}
\ifCLASSINFOpdf
  \usepackage[pdftex]{graphicx}
  \graphicspath{{figures}}
  \DeclareGraphicsExtensions{.pdf,.jpeg,.png}
\else
  \usepackage[dvips]{graphicx}
  \graphicspath{{figures}}
  \DeclareGraphicsExtensions{.eps}
\fi
\usepackage[cmex10]{amsmath}
\usepackage{amssymb}
\usepackage{amsthm}
\usepackage{bbm}
\usepackage{algorithmic}
\usepackage{array}
\usepackage{mdwmath}
\usepackage{mdwtab}
\usepackage{eqparbox}
\usepackage[caption=false,font=normalsize,labelfont=sf,textfont=sf]{subfig}
\usepackage{fixltx2e}
\ifCLASSOPTIONcaptionsoff
  \usepackage[nomarkers]{endfloat}
 \let\MYoriglatexcaption\caption
 \renewcommand{\caption}[2][\relax]{\MYoriglatexcaption[#2]{#2}}
\fi
\usepackage{url}
\usepackage{comment}
\usepackage{pgf,tikz,pgfplots}
\usetikzlibrary{arrows,automata,fit,positioning}
\usepackage{float}
\newfloat{Algorithm}{t}{lop}
\newtheorem{theorem}{Theorem}
\newtheorem{definition}[theorem]{Definition}
\newtheorem{lemma}[theorem]{Lemma}
\newtheorem{proposition}[theorem]{Proposition}
\renewcommand{\text}[1]{{\textnormal{#1}}}

\newcommand{\figref}[1]{{Fig.}~\ref{#1}}
\newcommand{\secref}[1]{{section}~\ref{#1}}

\newcommand{\tabref}[1]{{Table}~\ref{#1}}
\newcommand{\thref}[1]{{Theorem}~\ref{#1}}
\newcommand{\aref}[1]{{Algorithm}~\ref{#1}}
\newcommand{\propref}[1]{{Proposition}~\ref{#1}}
\newcommand{\lemref}[1]{{Lemma}~\ref{#1}}

\def\suchthat{\text{s.t.}}

\def\to{{\,\rightarrow\,}}
\def\kron{\otimes}

\let\avec\vec
\renewcommand{\vec}[1]{{\text{vec}(#1)}}
\newcommand{\indicator}[1]{\mathbbm{1}_{\left[ {#1} \right] }}

\def\half{\frac{1}{2}}
\newcommand{\fracl}[1]{\frac{1}{#1}} 
\providecommand{\abs}[1]{{\left\lvert#1\right\rvert}}
\providecommand{\trans}[1]{{#1^\dagger}}
\providecommand{\inv}[1]{{#1^{-1}}}
\providecommand{\inner}[2]{{\left\langle#1, #2\right\rangle}}
\providecommand{\innere}[2]{{\trans{#1}#2}}

\newcommand{\tr}{\mathrm{tr}}

\providecommand{\norm}[1]{{ \left\lVert#1\right\rVert }}
\providecommand{\normh}[2]{\norm{#1}_{#2}}
\newcommand{\normhs}[2]{\normh{#1}{#2}^2}
\newcommand{\normf}[1]{\normh{#1}{\text{F}}}
\newcommand{\normfs}[1]{\normf{#1}^2}
\newcommand{\normb}[1]{\normh{#1}{2}}
\newcommand{\normbs}[1]{\normb{#1}^2}

\newcommand{\vertiii}[1]{{\left\vert\kern-0.25ex\left\vert\kern-0.25ex\left\vert #1 
    \right\vert\kern-0.25ex\right\vert\kern-0.25ex\right\vert}}
\newcommand{\normt}[1]{\norm{#1}_{*}}
\newcommand{\normtb}[2]{\normh{#1}{#2, *}}
\newcommand{\normht}[2]{\normtb{#1}{#2}}

\newcommand{\gp}[2]{\cG\cP\left(#1, #2\right)}
\newcommand{\g}[2]{\cN\left(#1, #2\right)}

\newcommand{\D}[2]{\frac{d #1}{d #2}} 

\newcommand{\vect}[1]  {{\boldsymbol{#1}}}

\def\bphi{\vect{\phi}}
\def\bpsi{\vect{\psi}}

\def\bPsi{\vect{\Psi}}
\def\bPhi{\vect{\Phi}}

\def\bAlpha{\vect{\vA}}

\def\vc{{\vect{c}}}

\def\vg{{\vect{g}}}

\def\vl{{\vect{l}}}

\def\vr{{\vect{r}}}

\def\vu{{\vect{u}}}
\def\vv{{\vect{v}}}

\def\vx{{\vect{x}}}
\def\vy{{\vect{y}}}
\def\vz{{\vect{z}}}
\def\vA{{\vect{A}}}

\def\vW{{\vect{W}}}

\def\sc{{\avec{\vc}}}

\def\sg{{\avec{\vg}}}

\def\sl{{\avec{\vl}}}
\def\sm{{\avec{\vm}}}
\def\sn{{\avec{\vn}}}

\def\sr{{\avec{\vr}}}

\def\su{{\avec{\vu}}}
\def\sv{{\avec{\vv}}}

\def\sy{{\avec{\vy}}}

\def\bx{{\mathbf{x}}}
\def\by{{\mathbf{y}}}
\def\bz{{\mathbf{z}}}
\def\bA{{\mathbf{A}}}
\def\bB{{\mathbf{B}}}
\def\bC{{\mathbf{C}}}
\def\bD{{\mathbf{D}}}

\def\bG{{\mathbf{G}}}

\def\bI{{\mathbf{I}}}

\def\bK{{\mathbf{K}}}
\def\bL{{\mathbf{L}}}

\def\bP{{\mathbf{P}}}

\def\bR{{\mathbf{R}}}
\def\bS{{\mathbf{S}}}

\def\bU{{\mathbf{U}}}
\def\bV{{\mathbf{V}}}
\def\bW{{\mathbf{W}}}
\def\bX{{\mathbf{X}}}

\def\bZ{{\mathbf{Z}}}

\def\bbB{{\mathbb{B}}}

\def\bbE{{\mathbb{E}}}

\def\bbR{{\mathbb{R}}}

\def\cB{\mathcal{B}}

\def\cD{\mathcal{D}}

\def\cG{\mathcal{G}}
\def\cH{\mathcal{H}}

\def\cK{\mathcal{K}}

\def\cN{\mathcal{N}}
\def\cO{\mathcal{O}}
\def\cP{\mathcal{P}}

\def\sfD{\mathsf{D}}

\def\sfM{\mathsf{M}}
\def\sfN{\mathsf{N}}

\def\sfT{\mathsf{T}}

\newcommand{\compat}[2]{#1 \leadsto #2}
\newcommand{\sort}[1]{\text{sort}(#1)}
\newcommand{\mR}[1]{\bbR^{#1}_\downarrow}

\newcommand{\mB}[1]{\bbB^{#1}_\downarrow}
\newcommand{\Do}[1]{\Delta^{#1}}
\newcommand{\mDo}[1]{\Do{#1}_\downarrow}
\newcommand{\mDob}[1]{\mR{#1} \cap \Do{#1} }

\def\sm{_{\mbox{\tiny M}}}
\def\sn{_{\mbox{\tiny N}}}
\def\km{\bK_{\mbox{\tiny M}}}
\def\kn{\bK_{\mbox{\tiny N}}}

\def\gm{\bG_{\mbox{\tiny M}}}
\def\gn{\bG_{\mbox{\tiny N}}}
\def\xm{\bX_{\mbox{\tiny M}}}
\def\xn{\bX_{\mbox{\tiny N}}}

\def\ckm{\cK_{\mbox{\tiny M}}}
\def\ckn{\cK_{\mbox{\tiny N}}}
\def\cknm{\ckn\kron\ckm}

\def\kt{\bK_{\sfT}}

\def\mn{{m,n}}
\def\smn{_{m,n}}
\def\hm{\cH_{\ckm}}
\def\hn{\cH_{\ckn}}

\def\hk{\cH_{\cK}}

\def\data{\cD}

\def\sigmad{\sigma^2}

\def\dzero{\sfD^{-}}
\def\dones{\sfD^{+}}
\newcommand{\prtn}[3]  {\left[ \! \begin{smallmatrix}
                       \vect{#1}_{#2} \\   \hline
                       \vect{#1}_{#3}
                       \end{smallmatrix} \! 
                       \right] }
\newcommand{\prta}[1]  {\prtn{#1}{1}{2}}
\newcommand{\prtb}[1]  {\prtn{#1}{2}{1}}
\newcommand{\prtnb}[4]  {\left[ \! \begin{smallmatrix}
                       \vect{#1}^{#4}_{#2} \\   \hline
                       \vect{#1}^{#4}_{#3}
                       \end{smallmatrix} \! 
                       \right] }

\newcommand{\operatorat}[2]{\text{#1}_{@#2}}
\newcommand{\ap}[1]{\operatorat{AP}{#1}}
\newcommand{\recall}[1]{\operatorat{R}{#1}}
\newcommand{\pres}[1]{\operatorat{P}{#1}}

\def\auc{\text{AUC}}
\def\map{\text{MAP}}

\begin{document}
\title{The trace norm constrained matrix-variate Gaussian process for multitask
bipartite ranking}
\author{Oluwasanmi~Koyejo,~%
        Cheng~Lee,~%
        and~Joydeep~Ghosh,~\IEEEmembership{Fellow,~IEEE}%
\IEEEcompsocitemizethanks{\IEEEcompsocthanksitem O. Koyejo and J. Ghosh
are with the Department of Electrical and Computer Engineering, and C. Lee
is with the Department of Biomedical Engineering, University of Texas at Austin,
Austin, TX, 78712.\protect\\
E-mail: \{sanmi.k@, chlee@, ghosh@ece\}.utexas.edu}
\thanks{}}

\IEEEcompsoctitleabstractindextext{%
\begin{abstract}
We propose a novel hierarchical model for multitask bipartite ranking.
The proposed approach combines a matrix-variate Gaussian process with a
generative model for task-wise bipartite ranking. In addition, we employ a novel
trace constrained variational inference approach to impose low rank structure on the
posterior matrix-variate Gaussian process.
The resulting posterior covariance function is derived in closed form, and the
posterior mean function is the solution to a matrix-variate regression with a novel
spectral elastic net regularizer. Further, we show that variational inference
for the trace constrained matrix-variate Gaussian process combined with maximum
likelihood parameter estimation for the bipartite ranking model is jointly
convex.

Our motivating application is the prioritization of candidate disease genes.
The goal of this task is to aid the identification of unobserved associations
between human genes and diseases using a small set of observed associations as
well as kernels induced by gene-gene interaction networks and disease
ontologies.
Our experimental results illustrate the performance of the proposed model on
real world datasets.
Moreover, we find that the resulting low rank solution improves the
computational scalability of training and testing as compared to baseline models.
\end{abstract}

}

\maketitle

\IEEEdisplaynotcompsoctitleabstractindextext
\IEEEpeerreviewmaketitle

\ifCLASSOPTIONcompsoc
  \noindent\raisebox{2\baselineskip}[0pt][0pt]%
  {\parbox{\columnwidth}{\section{Introduction}\label{sec:introduction}%
  \global\everypar=\everypar}}%
  \vspace{-1\baselineskip}\vspace{-\parskip}\par
\else
  \section{Introduction}\label{sec:introduction}\par
\fi

\IEEEPARstart{R}{anking} is the task of learning an ordering for a set of items.
In bipartite ranking, these items are drawn from two sets, known as the positive
set and the negative set. Bipartite ranking involves learning an ordering that
ranks the positive items ahead of the negative items \cite{cortes04, agarwal05,
kotlowski11, gao12}.
This paper proposes a generative model for bipartite ranking and an extension of
bipartite ranking to the multitask domain. Our approach combines a latent
multitask regression function with task-wise ordered observation variables. We
employ a non-parametric matrix-variate Gaussian process prior for the multitask
regression. Further, we propose a novel trace constrained variational inference
approach that imposes useful low rank structure on the multitask regression.

Multitask learning (MTL) exploits inter-task relationships to improve the
prediction quality over single task learning\cite{evingou04, evingou05}. An
important class of methods in this domain are based on the matrix-variate
Gaussian process (MV-GP) and closely related models for vector valued
reproducing kernel Hilbert space (RKHS) function estimation \cite{alvarez11}.
The MV-GP is an extension of the matrix-variate Gaussian distribution
\cite{stegle11} to (possibly) infinite dimensional feature spaces.
Alternatively, the MV-GP may be understood as an extension of the scalar valued
Gaussian process \cite{rasmussen05} to vector valued responses. The MV-GP is a
useful model for learning multiple correlated tasks, as it jointly models the
correlations across examples, and across tasks.
The MV-GP has been applied to link analysis, transfer learning \cite{yu08},
collaborative prediction \cite{kai09} and multitask learning \cite{bonilla2008}
among other applications.

Our motivating application is the prioritization of disease genes. Genes are
segments of DNA that determine specific characteristics; over 20,000 genes have
been identified in humans, which interact to regulate various functions in the
body. Researchers have identified thousands of diseases, including various
cancers and respiratory diseases such as asthma \cite{gene11}, caused by
mutations in these genes.
The standard approach for discovering disease-causing genes are genetic
association studies \cite{mccarthy08}. However, these studies are often tedious
and expensive to conduct. Hence, computational methods that can reduce the
search space by predicting a prioritized list of candidate genes for a given
disease are of significant scientific interest.

The disease-gene prioritization task has received a significant amount of study
in recent years \cite{vanunu10, li10, mordelet11, natarajan11}. The task is
challenging because all the observed responses correspond to known associations
and the states of the unobserved associations are unknown, i.e., there are no
reliable negative examples. Such problems are also known \emph{single class} or
\emph{positive-unlabeled} (PU) learning tasks \cite{elkan08}.
A common approach for this task is to learn a model that that maximizes the
classification accuracy between the positive class and the unlabeled class
\cite{rendle09}. In the collaborative filtering literature, such single class
tasks have also been addressed using the low rank matrix factorization approach
\cite{pan08}.

Recent work suggests that a model trained to rank the positive class ahead of
the unknowns can be effective for ranking the unknown positive items ahead of
unknown negative items \cite{elkan08}. Further, the scientific use case for gene
prioritization depends on (and is evaluated by) the accuracy of the ranked list
produced \cite{vanunu10, mordelet11}. For these reasons, disease-gene
prioritization is well posed as a bipartite ranking task. A low rank model
induces significant correlation between the predictions of different tasks. This
assumption matches observations made by domain experts that the gene ranking
profiles of diseases are exhibit strong correlations \cite{sun11, natarajan11}.
This assumption is further validated by the empirical performance of the low
rank model.

Low rank structure is a typical constraint in several real world multitask and
matrix learning problems. However, despite its use for multitask learning, the
standard MV-GP does not model low rank structure. Further, memory requirements
for the MV-GP scale quadratically with data size, and na\"{i}ve computation scales
cubically with data size. These computational properties limit the applicability
of the MV-GP to large scale problems. The hierarchical factor Gaussian process
(factor GP) has been proposed as an alternative for problems with low rank
structure \cite{yu07, zhu09}. Here, latent row and column factors are drawn from
a Gaussian process prior. The result is a model with mean of user-selected rank.
We argue that the factor GP is an unsatisfactory model for two reasons:
(i) the resulting posterior mean function is the solution of a non-convex
optimization problem, and (ii) the posterior covariance is often intractable.
We will show that the proposed trace constrained MV-GP provides the same low
rank structure benefits without the drawbacks of the factor GP model. The
proposed variational inference is jointly convex in the mean and the covariance,
and the posterior covariance is given in closed form.

As a computational model, the optimization problem for the mean function of the
trace constrained MV-GP is equivalent to kernel multitask learning with the sum
of squared errors cost function \cite{koyejo11} combined with a novel
regularizer. We will show that this regularizer can be expressed as a weighted
sum of the Hilbert and the trace norms. We call the resulting regularization the
\emph{spectral elastic net}, highlighting its relationship to elastic net
regularization for variable selection in finite dimensional linear models
\cite{zou05}. To the best of our knowledge, ours is the first application of the
spectral elastic net regularizer to matrix estimation and kernel multitask
learning.

This paper proposes a novel generative model for multitask bipartite ranking and
a novel constrained variational inference approach for the matrix variate
Gaussian process applied to the disease-gene prioritization task. The main
contributions of this paper are as follows:
\begin{itemize}
  \item We propose a novel variational inference approach for matrix-variate
  Gaussian process regression using a trace norm constraint
  (\secref{sec:tracegp}). This constraint typically results in a regression matrix of low rank.
  \item We propose a novel generative model for bipartite ranking
  (\secref{sec:binary}).
  To our knowledge, ours is the first such generative model proposed in the
  literature.
  \item We show that variational inference for the latent regression model
  combined with maximum likelihood parameter estimation for the bipartite
  ranking is jointly convex (\secref{sec:inference_and_learning}).
  \item We evaluate the proposed model empirically and show that it outperforms
  the state of the art domain specific model for the disease-gene
  prioritization task (\secref{sec:experiments}).
\end{itemize}

{\bf The Kronecker product and the vec operator: } We will make significant use
of the Kronecker product and the $\vec{\cdot}$ operator. Given a matrix $\bA
\in \bbR^{P \times Q}$, $\vec{\bA} \in \bbR^{PQ}$ is the vector obtained by
concatenating columns of $\bA$. Given matrices $\bA \in \bbR^{P \times Q}$ and
$\bB \in \bbR^{P' \times Q'}$, the Kronecker product of $\bA$ and $\bB$ is
denoted as $\bA \kron \bB\in \bbR^{PP'\times QQ'}$. A useful property of the
Kronecker product is the identity: $\vec{\bA \bX \bB} = (\trans{\bB} \kron \bA)
\vec{\bX}$, where $\bX \in \bbR^{Q \times P'}$.

\section{The matrix-variate Gaussian process}\label{sec:mvgp}
The matrix-variate Gaussian process (MV-GP) is a collection of random variables
defined by their joint distribution for finite index sets. Let $\sfM \!\ni\! m$
be the set representing the rows (tasks) and $\sfN \!\ni\! n$ be the set
representing the columns (examples), with sizes $|\sfM|=M$ and $|\sfN|=N$. 
Let $X \sim \gp{\phi}{\cK}$, where $\gp{\phi}{\cK}$ denotes the MV-GP
with mean function $\phi$ and covariance function $\cK$. As with the scalar GP,
the MV-GP is completely specified by its mean function and its covariance
function. These are defined as:
\begin{align*}
&\phi(m,n) = \bbE[X(m,n)] \\
&\cK((m,n), (m',n')) =\\
&\quad \quad \quad \quad \quad \bbE[(X(m,n)-\phi(m,n))(X(m',n')-\phi(m',n'))],
\end{align*}
where $\bbE[\cdot]$ is the expected value. For a finite index set $\sfM
\times \sfN$, define the matrix $\bX \in \bbR^{M\times N}$ such that $x\smn =
X(m,n)$, then $\vec{\bX}$ is a distributed as a multivariate Gaussian with mean
$\vec{\bPhi}$ and covariance matrix $\bK$, i.e., $\vec{\bX}\sim
\g{\vec{\bPhi}}{\bK}$, where $\phi\smn = \phi(m,n)$, $\bPhi \in \bbR^{M\times
N}$, and $k_{(m,n), (m',n')}=\cK((m,n)(m', n'))$, $\bK \in \bbR^{MN \times MN}$.

The covariance function of the prior MV-GP is assumed to have Kronecker product
structure \cite{kai09, alvarez11}. The Kronecker product prior covariance
captures the assumption that the prior covariance between matrix entries can be
decomposed as the product of the row and column covariances. The Kronecker prior
covariance assumption is a useful restriction as: (i) it improves computational
tractability, enabling the model to scale to larger problems than may be
possible with a full joint prior covariance, (ii) the regularity imposed by the
separability assumption improves the reliability of parameter estimates even
with significant data sparsity, e.g.,
when the observed data consists of a single matrix \mbox{(sub-)sample}, and (iii)
row-wise and column-wise prior covariance functions are often the only prior
information available.
A closely related concept is kernel MTL with \emph{separable kernels}. This is a
special case of vector valued regularized function estimation where the joint
kernel decomposes into the product of the row kernel and the column kernel
\cite{evingou05, alvarez11, koyejo11}. Learning in these models is analogous to
inference for the MV-GP, with the prior row (resp.~column) covariance matrix
used as row (resp.~column) kernels.

Define the row covariance (kernel) function $\ckm: \sfM \times \sfM \mapsto
\bbR$ and the column covariance function $\ckn: \sfN \times \sfN \mapsto \bbR$.
The joint covariance function of the MV-GP with Kronecker covariance
decomposes into product form as $\cK\!\left((m,n),(m',n')\right) =
\ckm(m,m')\ckn(n,n')$, or equivalently, $\cK = \cknm$. Hence, for the random
variable $X \sim \gp{\phi}{\ckn \kron \ckm}$ and a finite index set $\sfM \times
\sfN$, $\vec{\bX} \sim \g{\vec{\bPhi}}{\kn \kron \km}$, where $\km \in \bbR^{M
\times M}$ is the row covariance matrix and and $\kn \in \bbR^{N \times N}$ is
the column covariance matrix.

This definition also extends to finite subsets that are not complete matrices.
Given any finite subset $\sfT\!=\!\{t\!=\!(m, n)\, |\, m\in \sfM, n \in \sfN
\}$, where  $T=|\sfT|\le M\times N$, the vector $\bx = \left[ x_{t_1} \ldots
x_{t_{T}} \right]$ is distributed as $\bx \sim \g{ \bPhi_{\sfT} }{\bK}$. The
vector $\bPhi_{\sfT} =[\phi(1) \ldots \phi(T)] \in \bbR^T$ are arranged from
the entries of the mean matrix corresponding to the set $t \in \sfT$, and $\bK$
is the covariance matrix evaluated only on pairs $t,t' \in \sfT \times \sfT$.

\begin{figure}[t]
\begin{center}
\begin{center}
\begin{tikzpicture}[semithick, node distance=4ex, label distance=-0.5ex, ->,
>=stealth', auto] 

\node[state, initial, initial right, initial distance=6ex, initial
text=$\cknm$] [] (z) {$Z$}; 

\node[state, initial, initial right,
initial distance=6ex, initial text=$\sigmad$, fill=black!25,
label=-60:$T$] [below =6.5ex of z] (r) {$r\smn$}; 

\node[draw, inner sep=2ex, rounded corners=1ex, fit=(r)] [] (plate-r)
{};

\path (z) edge (r);

\end{tikzpicture}
\end{center}
\end{center}
\caption{Plate model of the matrix-variate Gaussian process. $Z(\mn)$ is the
hidden noise-free matrix entry.}
\label{fig:mv-gp}
\end{figure}
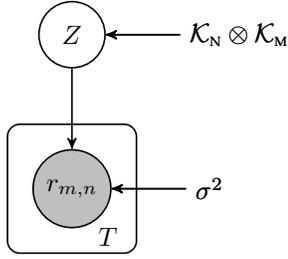

Our goal is to estimate an unknown response matrix $\bR \in \bbR^{M \times N}$
with $M$ rows and $N$ columns. We assume observed data consisting of a subset of
the matrix entries $\vr = \left[ r_{t_1} \ldots r_{t_{T}} \right]$ collected
into a vector. Note that $\sfT \subset \sfM \times \sfN$; hence, the data
represents a partially observed matrix.
Our generative assumption proceeds as follows (see \figref{fig:mv-gp}):
\begin{enumerate}
  \item Draw the function $Z$ from a zero mean MV-GP $Z \sim
  \gp{ {0} }{\cknm}$.
  \item Given $z\smn = Z(\mn)$, draw each observed response independently:
  $r\smn \! \sim \g{z\smn}{\sigmad}$.
\end{enumerate}
Hence, $\bZ \in \bbR^{M \times N}$ with entries $z\smn = Z(\mn)$ may be
interpreted as the latent noise-free matrix.
The inference task is to estimate the posterior distribution $Z|\data$,  where
$\data=\{\vr, \sfT\}$. The posterior distribution is again a Gaussian process,
i.e., $Z|\data \sim \gp{\phi}{\Sigma}$, with mean and covariance functions:
\begin{align} 
&\phi(m, n) = \kt(m,n)\inv{[\bK + \sigmad \bI_T ]}\vr \label{eq:post_mean}\\
&\Sigma\left((m, n), (m',n')\right) = \label{eq:post_var}\\
& \quad k((m,n),(m',n'))  - \kt(m,n)\inv{[\bK + \sigmad \bI_T ]}
\trans{\kt(m,n)}.\notag
\end{align}
where $\vr \in \bbR^T$ corresponds to the vector of responses for all training
data indexes $(m,n)\in \sfT$. The covariance function $\kt(m,n)$ corresponds to the
sampled covariance matrix between the index $(m,n)$ and all training data
indexes $(m',n')\in \sfT$, $\bK$ is the covariance matrix between all pairs
$(m,n),(m', n') \in \sfT \times \sfT$, and $\bI_T$ is the $T \times T$ identity
matrix.
The closed form follows directly from the definition of a MV-GP as a scalar
GP \cite{rasmussen05} with appropriately vectorized variables. The
model complexity scales with the number of observed samples $T$. Storing the
kernel matrix requires $\cO(T^2)$ memory, and the na\"{i}ve inference
implementation requires $\cO(T^3)$ computation.

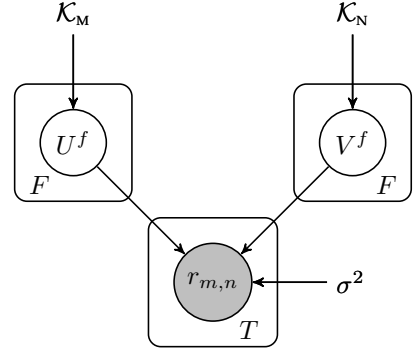
\begin{figure}[t]
\begin{center}
\begin{center}
\begin{tikzpicture}[semithick, node distance=4ex, label distance=-0.5ex, ->,
>=stealth', auto] 

\node[state, initial, initial above, initial distance=6ex, initial
text=$\ckm$, label=-120:$F$] [] (u) {$U^f$}; 

\node[state, initial, initial right,
initial distance=6ex, initial text=$\sigmad$, fill=black!25,
label=-60:$T$] [below right=10ex of u] (r) {$r\smn$}; 

\node[state, initial, initial above, initial distance=6ex, initial
text=$\ckn$, label=-60:$F$] [above right=10ex of r] (v)
{$V^f$};

\node[draw, inner sep=2ex, rounded corners=1ex, fit=(u)] [] (plate-u)
{};

\node[draw, inner sep=2ex, rounded corners=1ex, fit=(v)] [] (plate-v)
{};

\node[draw, inner sep=2ex, rounded corners=1ex, fit=(r)] [] (plate-r)
{};

\path (u) edge (r);
\path (v) edge (r);

\end{tikzpicture}
\end{center}
\end{center}
\caption{Plate model of the factor Gaussian process.}
\label{fig:factor-gp}
\end{figure}

Although the matrix-variate Gaussian process approach results in closed form
inference, it does not model in low rank matrix structure.
The factor GP is a hierarchical Gaussian process model that attempts to address
this deficiency in the MV-GP \cite{zhu09, yu07}.
With a fixed model rank $F$, the generative model for the factor GP is as
follows (see \figref{fig:factor-gp}):
\begin{enumerate}
  \item For each $f\in \{1\ldots F\}$, draw row functions: $U^f \sim \gp{ {0}
  }{\ckm}$. Let $\vu_m \in \bbR^F$ with entries $u^f_m = U^f(m)$.
  \item For each $f\in \{1\ldots F\}$, draw column functions: $V^f \sim \gp{
  {0} }{\ckn}$. Let $\vv_n \in \bbR^F$ with $v^f_n
  = V^f(n)$.
  \item Draw each matrix entry independently: $r\smn \! \sim
  \g{\innere{\vu_m}{\vv_n}}{\sigmad} \; \forall \,(m,n)\in \sfT$.
\end{enumerate}
where $\vu_m$ is the $m^{\text{th}}$ row of $\bU=[\vu^1 \ldots \vu^F]\in \bbR^{M\times
F}$, and $\vv_n$ is the $n^{\text{th}}$ row of $\bV=[\vv^1 \ldots \vv^F]\in
\bbR^{N\times F}$.
The maximum-a-posteriori (MAP) estimates of $\bU$ and $\bV$ can be computed as
the solution of the following optimization problem:
\begin{multline}\label{eq:factor_gp}
\bU^*,\bV^* = \underset{\bU, \bV}{\arg\min} \fracl{\sigmad}\sum_{(\mn)\in T}
(r\smn-\innere{\vu_m}{\vv_n})^2 \\
+ \tr(\trans{\bU}\inv{\km}\bU) + \tr(\trans{\bV}\inv{\kn}\bV)
\end{multline}
where $\tr(\bX)$ is the trace of the matrix $\bX$. However, the joint posterior
distribution of $\{\bU, \bV\}$ and the distribution of $\bZ=\bU\trans{\bV}$ are 
not Gaussian, and
the required expectations and posterior distributions are quite challenging
to characterize. A Laplace approximation by proposed by \cite{yu07} and
\cite{zhu09} utilized sampling techniques.
Statistically, the factor GP may be interpreted as the sum of rank-one factor
matrices. Hence, as the rank $F \to \infty$, the law of large numbers can be
used to show that the distribution of $Z$ converges to $\gp{0}{\ckn\kron\ckm}$
\cite{yu07}. 
\subsection{Spectral norms of compact operators}\label{sec:tracenorm}
The mean function of the MV-GP is an element of the Hilbert space defined by the
kernels (covariances). We provide a brief overview of some relevant background
required for defining this representation and for defining relevant spectral
norms of compact operators. We will focus on the MV-GP with Kronecker prior
covariance. Our exposition is closely related to the approach outlined in
\cite{abernethy09}. Further details may be found in \cite{berlinet07}.

Let $\hm$ denote the Hilbert space of functions induced by the row kernel
$\ckm$. Similarly, let $\hn$ denote the Hilbert space of  functions induced by
the column kernel $\ckn$. Let $\vx \in \hm$ and $\vy \in \hn$ define (possibly
infinite dimensional) feature vectors.
The mean function the MV-GP is defined by a linear map $W: \hm \mapsto \hn$, i.e.,
the bilinear form on $\hk=\hm \times \hn$ given by:
\begin{equation*}
\phi(m,n) = \inner{\vx_m}{W \vy_n}_{\hm}
\end{equation*}
Let $\cB$ denote the set of compact bilinear operators mapping $\hm \mapsto
\hn$. A compact linear operator $W \in \cB$ admits a spectral decomposition
\cite{abernethy09} with singular values given by $\{\xi_i(W)\}$. 

{\bf The trace norm} is given by the \emph{ell}-1 norm on the spectrum of $W$:
\begin{equation} \label{eq:trace_norm}
\normht{\phi}{\hk} = \sum_{i=1}^D \xi_i(W)
\end{equation}
When the dimensions are finite, \eqref{eq:trace_norm} is the trace norm
of the matrix $\vW \in \bbR^{D\sm \times D\sn}$. This norm has been widely
applied to several machine learning tasks including multitask learning
\cite{pong2010, dudik12} and recommender systems \cite{srebro05b}.
In addition to the trace norm, a common regularizer is the induced {\bf Hilbert
norm} given by the \emph{ell}-2 norm on the spectrum
of $W$:
\begin{equation} \label{eq:hilbert_norm}
\normhs{\phi}{\hk} = \sum_{i=1}^D \xi_i^2(W)
\end{equation}
\eqref{eq:hilbert_norm} is equivalent to the matrix
Frobenius norm for finite dimensional $\bW \in \bbR^{D\sm \times D\sn}$ computed
as: 
\begin{equation*}
\normf{\bW}^2 = \sum_{i=1}^{\min(D\sm, D\sn)} \xi_i^2(\bW)
\end{equation*}

Let $L(\phi, \vr, \sfT)$ represent the loss function for a finite set of
training data points $\sfT \in \sfM \times \sfN$ and $Q(\phi)$ be a spectral
regularizer. We define the regularized risk functional:
\begin{equation*}
L(\phi, \vr, \sfT) + \lambda Q(\phi)
\end{equation*}
where $\lambda \ge 0$ is the regularization constant. A representer theorem
exists, i.e., the function $\phi$ that optimizes the
regularized risk can be represented as a finite weighted
sum of the kernel functions evaluated on training data\cite{abernethy09}.
Employing this representer theorem, the optimizing function can be computed as:
\begin{align}\label{eq:representer}
\phi(m,n) 
&=\sum_{m' \in \sfM} \sum_{n' \in \sfN} \alpha_{m',n'} \,\ckm(m, m')\ckn(n,
n')\notag\\
&=\km(m)\bAlpha\trans{\km(n)}
\end{align}
where $\bAlpha \in \bbR^{M \times N}$ is a parameter matrix, $\km(m)$ is the kernel
matrix evaluated between $m$ and $m' \in \sfM$, i.e., the $m^{\text{th}}$ row of $\km$, and
$\kn(n)$ is the kernel matrix evaluated between $n$ and all $n' \in \sfN$. 

{\bf Computing the norms:} The Hilbert norm can be computed as:
\begin{align}\label{eq;hilber_norm}
\normhs{\phi}{\hk}
&=\trans{\vec{\bAlpha}}(\kn \kron \km)\vec{\bAlpha} \notag\\
&=\tr(\trans{\bAlpha}\km \bAlpha\kn).
\end{align}
The trace norm can be computed using a basis transformation
approach\cite{abernethy09} or by using the low rank ``variational''
approximation \cite{abernethy09, dudik12}. 

{\bf Basis transformation:} With the index set fixed, define bases $\gm\in
\bbR^{M \times D\sm}$ and $\gn\in \bbR^{N \times D\sn}$ such that $\km =
\gm\trans{\gm}$ and $\kn = \gn\trans{\gn}$. One such basis is the square root of
the kernel matrix $\gm=\km^\half$ and $\gn=\kn^\half$. When the feature space is
finite dimensional, the feature matrices $\xm \in \bbR^{M \times D\sm}$ and $\xn
\in \bbR^{N \times D\sn}$ are also an appropriate basis. The mean function can
be re-parametrized as $\phi(m,n) = \gm(m)\bB\trans{\gn(n)}$, where $\bB \in
\bbR^{D\sm \times D\sn}$. Now, the trace norm is given directly by the trace
norm of the parameter matrix, i.e., $\normht{\phi}{\hk}=\normt{\bB}$.

{\bf Low rank ``variational'' approximation:} The trace norm can also be
computed using the low rank approximation. This is sometimes known as the
variational approximation of the trace norm \cite{dudik12}.
\begin{equation}\label{eq:low_rank} 
	\normht{\phi}{\hk} = \underset{\phi =
	\inner{u}{v} }{\arg\min} \; \half\sum_{f=1}^F \left(\normhs{u^f}{\hm} +
	\normhs{v^f}{\hn} \right)
\end{equation} 
where $\inner{u}{v}=\sum_{f=1}^F u^f v^f$. This approach is exact when $F$
is larger than the true rank of $\phi$. Note that this is the same regularization
that is required for MAP inference with the factor GP model
\eqref{eq:factor_gp}. Hence, when $F$ is sufficiently large, the regularizer in 
the factor GP model is the trace norm. Unfortunately, it is difficult to select
an appropriate rank a-priori, and no such claims exist when $F$ is
insufficiently large. With finite dimensions, the variational approximation of
the trace norm is given by:
\begin{equation*} 
	\normt{\bW} = \underset{\bW =
	\bU\trans{\bV} }{\arg\min} \; \half \left(\normf{\bU}^2 +
	\normf{\bV}^2 \right)
\end{equation*}
where $\bU \in \bbR^{D\sm \times F}$ and $\bV \in \bbR^{D\sn \times F}.$ This sum
of factor norms has proven effective for the regularization of matrix
factorization models and other latent factor problems \cite{ruslan08}.

\section{Trace norm constrained inference for the MV-GP}\label{sec:tracegp}
A generative model for low rank matrices has proven to be a challenging problem.
We are unaware of any (non-hierarchical) distributions in the literature that
generate sample matrices of low rank. Hierarchical models have been proposed, but
such models introduce issues such as non-convexity and non-identifiability of
parameter estimates. Instead of seeking a generative model for low rank matrices, 
we propose a variational inference approach. We constrain the inference of the MV-GP 
such that expected value of the approximate posterior distribution has a constrained 
trace norm (and is generally of low rank). In contrast to standard variational 
inference, this constraint is enforced in order to extract structure.

The goal of inference is to estimate of the posterior distribution $p(Z|\data)$.
We propose approximate inference using the log likelihood lower bound
\cite{bishop06}:
\begin{equation}\label{eq:var_bnd}
\ln p(\by|\data) \ge \bbE_{\bZ}[\ln p(\by, \bZ)] - \bbE_{\bZ}[\ln p(\bZ)]
\end{equation}
Our approach is to restrict the search to the space of Gaussian processes
$q(Z)=\gp{\psi}{S}$ subject to a trace norm constraint $\normht{\psi}{\cK} \le
C$ as defined in \eqref{eq:trace_norm}. With no loss of generality, we
assume a set of rows $\sfM$ and columns $\sfN$ of interest so $\sfT \in \sfM
\times \sfN$. Let $\bZ \in \bbR^{M \times N}$ be the matrix of hidden
variables.

Given finite indexes, $q$ is a Gaussian distribution $q(\bz) = \g{\bpsi}{\bS}$
where $\bz = \vec{\bZ} \in \bbR^{M \times N}$, $\bphi = \vec{\bPsi} \in \bbR^{M
\times N}$, and $\bS\in \bbR^{MN \times MN}$.
Let $\bP \in \bbR^{T \times MN}$ be a  permutation matrix such that $\bS_{T} =
\bP\bS\trans{\bP}$ is the covariance matrix of the subset of observed entries $t
\in \sfT$, and $\bK_{T} = \bP\bK\trans{\bP}$ is the prior covariance of the
corresponding subset of entries.
Evaluating expectations, the lower bound \eqref{eq:var_bnd} results in the
following inference cost function (omitting terms independent of $\bpsi$ and
$\bS$):
\begin{gather*}
\underset{\bpsi, \bS}{\max} \; 
-\fracl{2\sigmad} \sum_{\mn \in
\sfT}(r\smn-\psi\smn)^2 -\fracl{2\sigmad} \tr( \bS_{T}) \notag \\
-\half \trans{\bpsi}\inv{\bK}\bpsi - \half
\tr(\inv{\bK}\bS) + \ln|\bS| \notag\\
\quad \quad \text{s.t.}\; \normht{\psi}{\cK} \le C 
\label{eq:all_cost}
\end{gather*}
where $|\bX|$ is the determinant of matrix $\bX$.

First, we compute gradients with respect to $\bS$. After setting the gradients
to zero, we compute:
\begin{align}\label{eq:qsigma}
\bS^* &= \inv{\left({ \inv{\bK} + \fracl{\sigmad} \trans{\bP}\bP
}\right)} \notag \\
&= \bK - \bK\trans{\bP}\inv{\left({\bK_T +
\fracl{\sigmad} \bI_T }\right)} \bP\bK,
\end{align}
The second equality is a consequence of the matrix inversion lemma.
Interestingly, \eqref{eq:qsigma} is identical the posterior
covariance of the unconstrained MV-GP \eqref{eq:post_var}. 

Next, we collect the terms involving the mean. This results in the optimization
problem:
\begin{gather}
\bpsi^*= \underset{\bpsi }{\arg\min} \;\fracl{2\sigmad} \sum_{m,n \in
\sfT}(r\smn-\psi\smn)^2 +\half \trans{\bpsi}\inv{\bK}\bpsi \notag \\
\quad \quad \text{s.t.}\; \normht{\psi}{\hk} \le C \label{eq:qmean}
\end{gather}
This is a convex regularized least squares problem with a convex
constraint set. Hence, \eqref{eq:qmean} is convex, and $\bpsi^*$ is unique.
Using the Kronecker identity, we can re-write the cost function in parameter
matrix form. We can also replace the trace constraint with the equivalent
regularizer weighed by $\mu$. Multiplying through by $\sigmad$ leads to
the equivalent cost:
\begin{align}\label{eq:qmean_reg}
\bPsi^*
&=\underset{\Psi }{\arg \min} \; \half \sum_{m,n \in
\sfT}(r\smn-\psi \smn)^2 \notag\\
&\quad+\frac{\sigmad}{2} \tr(\trans{\bPsi}\inv{\km}\bPsi\inv{\kn})
+\mu\sigmad \normht{\psi}{\hk},
\end{align}

Applying the representation \eqref{eq:representer}, we recover the
parametric form of the mean function $\psi \in \cknm$ as $\bPsi = \km \bAlpha
\kn$ where $\bAlpha \in \bbR^{M \times N}$. We may also solve for $\bAlpha$
directly:
\begin{align}\label{eq:qmean_kernel}
\bAlpha^*
&=\underset{ \bAlpha }{\arg \min} \; \half \sum_{\mn \in
\sfT} \left(r\smn-(\km\bAlpha\kn)\smn \right)^2 \notag\\
&\quad+\frac{\sigmad}{2} \tr(\trans{\bAlpha}\km\bAlpha \kn)
+\mu\sigmad \normht{\psi(\bAlpha)}{\hk}.
\end{align}
where $\psi(\bAlpha)$ is the mean function corresponding to
the parameter $\bAlpha$ (see \eqref{eq:representer}). The
representation of the mean function in functional form is useful for avoiding
repeated optimization when testing a trained model with different evaluation
sets.

The approximate posterior distribution is itself a finite index set
representation of an underlying Gaussian process.
\begin{theorem}\label{thrm:gp}
The posterior distribution $q=\g{\bpsi}{\bS}$ is the finite index set
representation of the Gaussian process $\gp{\psi}{S}$ 
where the mean function $\psi$ is given by \eqref{eq:qmean_kernel} and the
covariance function $S$ is given by \eqref{eq:post_var}.
$g=\gp{\psi}{S}$ is the unique posterior distribution
that maximizes the lower bound of the log likelihood \eqref{eq:var_bnd}
subject to the trace constraint $\normht{\psi}{\cK} \le C$.
\end{theorem}
{\bf Sketch of proof:}
Uniqueness of the solution follows from \eqref{eq:var_bnd}, which is jointly convex
in $\{\bpsi, \bS\}$. To show that the posterior distribution is a Gaussian
process, we simply need to show that for a fixed training set $\data$, the
posterior distribution of the superset $(\sfM \times \sfN)\cup (m',n')$ has
the same mean function and covariance function. These follow directly from the
solution of \eqref{eq:qmean_kernel} and from \eqref{eq:post_var} (see
\eqref{eq:qsigma}). In addition to showing uniqueness,
\thref{thrm:gp} shows how the trained model can be extended to evaluate the
posterior distribution of data points not in training.

In the case where a basis for $\km$ and $\kn$ can be found, \eqref{eq:qmean}
may be solved using the matrix trace norm approach directly
(see \secref{sec:tracenorm}):
\begin{align}\label{eq:qmean_matrix}
\bB^*=\underset{ \bB }{\arg \min} \; 
&\half \sum_{\mn \in
\sfT}\left(r\smn-(\gm\bB\trans{\gn})\smn \right)^2 \notag\\
&\quad+\frac{\sigmad}{2} \normfs{\bB}+ \mu\sigmad \normt{\bB}.
\end{align}
where $\gm\in \bbR^{M \times D\sm}$ is the basis for $\km$ and $\gn\in \bbR^{N
\times D\sn}$ is the basis for $\kn$. $\bB \in \bbR^{D\sm \times D\sn}$ is the
estimated parameter matrix. The mean function is then given by
$\psi\smn = (\gm\bB\trans{\gn})\smn$.

{\bf Spectral elastic net regularization:} The regularization that results from
the constrained inference has an interesting interpretation as the spectral
elastic net norm. As discussed in \secref{sec:tracenorm}, the mean function may
be represented as $\psi(m,n) = \inner{\vx_m}{W \vy_n}_{\hm}$ for $\vx \in \hm$
and $\vy \in \hn$. The spectral elastic net is given as a weighed sum of the
\emph{ell}-2 norm and the \emph{ell}-1 norms on the spectrum $\{\xi_i(W)\}$:
\begin{equation} \label{eq:enet_norm}
Q_{\alpha, \beta}(\psi) = \alpha \sum_{i=1}^D \xi_i^2(W) +
\beta \sum_{i=1}^D \xi_i(W)
\end{equation}
where $\alpha$ and $\beta \ge 0$ are weighting constants.
The naming is intentionally suggestive of the analogy to the elastic
net regularizer \cite{zou05}. The elastic net regularizer is a
weighted sum of the \emph{ell}-2 norm and the \emph{ell}-1 norms of the
parameter vector in a linear model. The elastic net is a tradeoff between
smoothness, encouraged by the \emph{ell}-2 norm, and sparsity, encouraged by the
\emph{ell}-1 norm. The elastic net is particularly useful when learning with
correlated features. The spectral elastic net has similar properties. The
Hilbert norm encourages smoothness over the spectrum, while the trace norm
encourages spectral sparsity, i.e., low rank. To the best of our knowledge, this
combination of norms is novel, both in the matrix estimation literature and in
the kernelized MTL literature. When the dimensions are finite,
\eqref{eq:enet_norm} is given by a weighted sum of the trace norm and the 
Frobenius norm of the parameter matrix. 

We propose a parametrization of the mean function inference inspired by the
elastic net \cite{zou05}. Let $\lambda = \sigmad(1+\mu)$ and
$\alpha=\frac{\mu\sigmad}{\sigmad(1+\mu)}$ where $\lambda\ge0$ and $\alpha \in
[0, 1]$. The loss function \eqref{eq:qmean_matrix} can be parametrized as:
\begin{align}\label{eq:qmean_reparam}
\bB^*=\underset{ \bB }{\arg \min} \; 
&\half \sum_{\mn \in \sfT}\left(r\smn-(\gm\bB\trans{\gn})\smn \right)^2
\notag\\
&\quad+\frac{\lambda(1-\alpha)}{2} \normfs{\bB}+ \lambda\alpha \normt{\bB}.
\end{align}
The same parametrization can also be applied to the equivalent representations
given in \eqref{eq:qmean_reg} and \eqref{eq:qmean_kernel}.
This spectral elastic net parametrization clarifies the tradeoff between the trace norm and the Hilbert norm. 
The trace norm is recovered when $\alpha=1$, and the Hilbert norm is recovered for $\alpha=0$. The
spectral elastic net approach is also useful for speeding up the computation with
warm-start i.e. for a fixed $\alpha$, we may employ warm-start for decreasing
values of $\lambda$.
Computation of the spectral elastic net
norm follows directly from the Hilbert and trace norms.
From the variational approximation of the trace norm \eqref{eq:low_rank}, it is
clear that MAP inference for the factor GP \eqref{eq:factor_gp} is equivalent to 
inference for the mean of the trace constrained MV-GP \eqref{eq:qmean_reg} in
the special case where $\alpha = 1$ (assuming that the non-convex optimization
\eqref{eq:factor_gp} achieves the global maximum).

{\bf Non-zero mean prior:} To simplify the explanation, we have assumed so far
that the prior Gaussian process has a zero mean. The non-zero mean case is a
straightforward extension \cite{rasmussen05}. We include a short discussion for
completeness. Let $b\smn$ represent the mean parameter of the Gaussian process
prior, i.e., $Z \sim \gp{b}{\cknm}$. The posterior covariance estimate remains the
same, and the posterior mean computation requires the same optimization, but
with the observation $r\smn$ replaced by $\tilde{r}\smn = r\smn - b\smn$. The
resulting posterior mean must then be shifted by the bias, i.e.,
$\bbE[Z\smn|\data] = \psi\smn + b\smn$.
If desired, this parameter may be easily estimated. Suppose we choose to model a
row-wise bias. Let $\sfT_m = \{(m, n) | (m', n) \in T, \,m'=m\}$, then solving
the straightforward optimization, we find that the the row bias estimate is given
by:
\begin{equation*}
b_m = \fracl{|\sfT_m|}\sum_{n | \mn \in T_m} r\smn - \psi\smn.
\end{equation*}

\section{Bipartite ranking}\label{sec:binary}
Bipartite ranking is the task of learning an ordering for items drawn from two
sets, known as the positive set and the negative set, such that the items in the
positive set are ranked ahead of the items in the negative set \cite{cortes04,
agarwal05, kotlowski11, gao12}. Many models for bipartite ranking attempt to
optimize the pair-wise mis-classification cost, i.e., the model is penalized for
each pair of data points where the positive labeled item is ranked lower than
the negative labeled item. Although this approach has proven effective, the
required computation is quadratic in the number of items. This quadratic
computation cost limits the applicability of pair-wise bipartite ranking to
large scale problems.

More recently, researchers have shown that it may be sufficient to optimize a
classification loss, such as the exponential loss or the logistic loss, directly
to solve the bipartite ranking problem \cite{kotlowski11, gao12}. This is also
known as the point-wise approach in the ranking literature.
In contrast to the point-wise and pair-wise approach, we propose a \emph{list-wise} bipartite
ranking model. The list-wise approach learns a ranking model for the entire set
of items and has gained prominence in the learning to rank literature
\cite{pradeep11, acharyya12} as it comes with strong theoretical guarantees and
has been shown to have superior empirical performance.

\begin{figure}[t]
\begin{center}
\includegraphics[width=0.8\columnwidth]{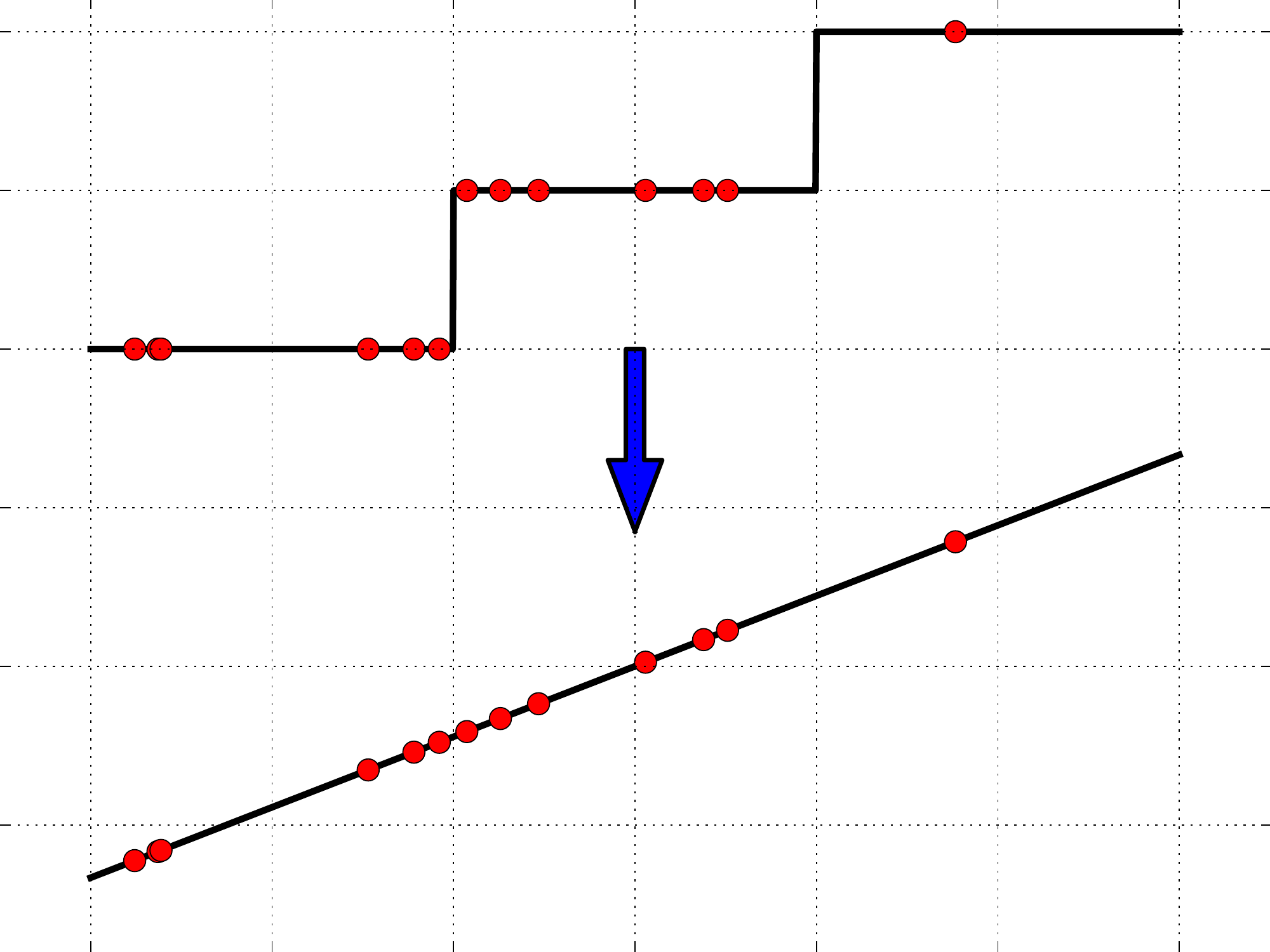}
\end{center}
\caption{Monotone re-targeting searches for an order preserving transformation
of the target scores that may be easier for the regressor to fit.}
\label{fig:mr}
\end{figure}
Our approach is inspired by monotone retargeting (MR) \cite{acharyya12}, a
recent method for adapting regression to ranking tasks. Although many
ranking models are trained to predict the relevance scores, there is no need to
fit scores exactly. Any scores that induce the correct ordering will suffice.
MR jointly optimizes over monotonic transformations of the target scores and an
underlying regression function (see \figref{fig:mr}).  We will show that
maximum likelihood parameter estimation in the proposed model is equivalent
to learning the target scores in MR. Though we show this equivalence for the
special case of bipartite ranking with square loss, the relation
holds for more general Bregman divergences and ranking tasks. This extension is
beyond the scope of this paper. In addition to improving performance, MR has
favorable statistical and optimization theoretic properties, particularly when
combined with a Bregman divergence such as squared loss. To the best of our
knowledge, ours is the first generative model for list-wise bipartite ranking.

\subsection{Background}\label{sec:background}
Let $\bbB = \{ +1, -1\}$, and let $\mB{d}$ be the set of binary isotonic vectors
(binary vectors in sorted order), i.e., any $\vv \in \mB{d}$ satisfies $\vv \in
\bbB^d$ and $v_i \ge v_j \; \forall j > i$. Similarly, let $\mR{d}$ be the set
of real valued isotonic vectors, i.e., any $\vv \in \mR{d}$ satisfies $\vv \in
\bbR^{d}$ and $v_i \ge v_j \; \forall j > i$, then $\vv$ satisfies
\emph{partial} order. We state that $\vv$ satisfies \emph{total} order or
\emph{strict isotonicity} when the ordering is a strict inequality, i.e.,
$v_i > v_j \; \forall j > i$.  We denote a vector in sorted order as $\sv =
\sort{\vv}$.

Compatibility is a useful concept for capturing the match between the sorted
order of two vectors.
\begin{definition}[Compatibility \cite{pradeep11}]\label{def:compatilibity}
$\vu$ is compatible with the sorted order of $\sv$ (denoted as
$\compat{\vu}{\sv}$) if for every pair of indexes $(i,j)$, $\sv_{i} \ge
\sv_{j}$ implies $\vu_{i} \ge \vu_{j}$.
\end{definition}
Compatibility is an asymmetric relationship, i.e., $\compat{\vu}{\sv}
\nRightarrow \compat{\sv}{\vu}$. It follows that sorted vectors always satisfy
compatibility, i.e., if $\su$ and $\sv$ are two sorted vectors, then by
definition \ref{def:compatilibity}, $\compat{\su}{\sv}$.
Compatibility is straightforward to check when the target vector is binary. Let
$\vu \in \bbR^d$ and $\sv \in \mB{d}$, and let $k$ be the be the number of
$+1$'s in the sorted vector $\sv$, i.e., the threshold for transition between $+1$
and $-1$. Then $\compat{\vu}{\sv}$ implies that:
\begin{equation}\label{eq:compat_binary} 
\exists \,b\in \bbR \quad
\suchthat \quad \underset{1\le i \le k}{\min} u_i 
\ge b \ge \underset{k< j \le d}{\max}u_j.
\end{equation}

There are several ways to permute a sorted binary vector $\sy \in \mB{d}$ while
keeping all its values the same. These are permutations that separately re-order
the $+1$s at the top of $\sy_m$ and the $-1$s at the bottom. Given $\sy =
\sort{\vy}$, we represent the set of permutations that do not change the value
the sorted $\sy$ as $\Gamma = \{ \gamma(\cdot) \; \vert \;\gamma(\sy) = \sy
\}$. It follows that the set $\Gamma$ contains all permutations that
satisfy $\compat{\gamma(\sv)}{\sy}$. In other words, all $\vv$ that satisfy
$\compat{\vv}{\sy}$ can be represented as $\vv = \gamma(\sv)$ for some
$\gamma \in \Gamma$.

We propose a representation for compatible vectors that reduces to permutations
of isotonic vectors.
\begin{proposition}\label{prep:represent}
Let $\sv \in \mB{d}$. Any $\vu \in \bbR^d$ that
satisfies $\compat{\vu}{\sv}$ can be represented by $\vu = \gamma(\su)$ where
$\gamma\in \Gamma$, the set $ \Gamma= \{ \gamma(\cdot) \; \vert \;\gamma(\sv) =
\sv \}$ and $\su \in \mR{d}$.
\end{proposition}
{\bf Sketch of proof:} 
First, we note that by definition of compatibility for binary vectors
\eqref{eq:compat_binary}, any permutation $\gamma\in \Gamma$ satisfies
$\compat{\gamma(\su)}{\sv}$. Next, we note that the sorted order is a member of
the permutation set. This representation is unique when $\su$ satisfies strict ordering.

The set $\mR{d}$ is a convex cone. To see this, note that the convex composition
$\vx = \alpha \vu + (1-\alpha) \vv, \, \alpha \in [0, 1] $ of two isotonic
vectors $\vu \in \mR{d}$ and $\vv \in \mR{d}$ preserves isotonicity. Further,
any scaling $\alpha \vx$ where $\alpha >0$ preserves the ordering.
Let $\Do{d}$ be the set of probability distributions, i.e., $\forall \, \vx \in
\Do{d}$, $x_i \ge 0$ and $\sum_{i=1}^d x_i = 1$. The set of probability
distributions in sorted order is given by $\mDo{d} = \mDob{d}$ so for each $\vx
\in \mDo{d}$, $\vx \in \Do{d}$ and $x_i \ge x_j\, \forall i > j$.
\begin{lemma}[Representation of $\mDo{d}$
\cite{acharyya12}]\label{lem:order_simplex} The set $\mDo{d}$ of all discrete
probability distributions of dimension $d$ that are in descending order is the
image $\bC\vx$ where $\vx \in \Do{d}$ and $\bC$ is an upper triangular
matrix generated from the vector $\vv = \{1, \frac{1}{2} \cdots \frac{1}{T}\}$
such that $\bC(i,:)= \{0\}^{i-1} \times \vv(i:)$
\end{lemma}

\subsection{Generative model}
Let $y_{m, n} \in \bbB$ be the label for item $n$ in $m^{\text{th}}$ task and
let $\sfT_m = \{n\, | \, (m, n) \in \sfT\}$ be the set of items in $m^{\text{th}}$ task so $|\sfT_m|
= T_m$. We define the negative set as $\dzero = \{(m, n) \in \sfT \,| \, y\smn
=-1 \}$ and the positive set as $\dones = \{(m, n) \in \sfT \,| \, y\smn =+1
\}$. For the $m^{\text{th}}$ task, the negative set is defined as $\dzero_m = \{n | (m,
n) \in \dzero \}$ and the positive set as $\dones_m = \{n | (m, n) \in \dones
\}$ so that $\sfT_m = \dones_m \cup \dzero_m$. The vector of labels for the
$m^{\text{th}}$ task are given by $\vy_m \in \bbB^{T_m}$.
We propose the following generative model for
$\vy_m$:
\begin{equation}\label{eq:generative}
p(\vy_m | \vr_m) \propto\prod_{l \in \dones_m}
\prod_{l' \in \dzero_m} \indicator{r_{m, l} \ge r_{m, l'}}.
\end{equation} %
where $\indicator{\cdot}$ is the indicator function
defined as:
\begin{equation*}
\indicator{b}
=\begin{cases}
1& \text{if $b$  evaluates to true },\\
0& \text{otherwise}.
\end{cases}
\end{equation*}
For clarity, we have suppressed the dependence of $p(\vy_m | \vr_m)$ on the sets
$\{ \dones_m, \dzero_m\}$.

It is instructive to compare the form of the generative model
\eqref{eq:generative} to the area under the ROC curve (AUC) given by the
fraction of correctly ordered pairs:
\begin{equation}\label{eq:auc}
\fracl{\abs{\dones_m}\abs{\dzero_m}}
\sum_{l \in \dones_m}
\sum_{l' \in \dzero_m} 
\left(\indicator{r_{m, l} > r_{m, l'}}
+ \half \indicator{r_{m, l} = r_{m, l'}}\right).
\end{equation}
We note that $p(\vy_m | \vr_m)$ is nonzero if and only if $\vr_m$ satisfies the
ordering defined by $\{ \dones_m, \dzero_m\}$. It follows that any vector $\vr_m \;
\suchthat \; p(\vy_m | \vr_m)$ is nonzero also maximizes the AUC.

\begin{figure}[t]
\begin{center}
\begin{tikzpicture}[semithick, node distance=4ex, label distance=-0.5ex, ->,
>=stealth', auto] 

\node[state, initial, initial right, initial distance=7ex, initial
text=$\cknm$] [] (z) {$Z$}; 

\node[state, initial, initial right,
initial distance=7ex, initial text=$\sigmad$, 
label=-70:$N_m$] [below =6.5ex of z] (r) {$r\smn$}; 

\node[state, initial, initial right,
initial distance=7ex, initial text=${\{ \dones, \dzero \}}$ , fill=black!25,
label=-50:$M$] [below =6.5ex of r] (y) {$\vy_m$}; 

\node[draw, inner sep=2.5ex, rounded corners=1ex, fit=(r)] [] (plate-r)
{};

\node[draw, inner sep=3.5ex, rounded corners=1ex, fit=(r)(y)] [] (plate-y)
{};

\path (z) edge (r);
\path (r) edge (y);

\end{tikzpicture}
\end{center}
\caption{Plate model of generative bipartite ranking with the latent
matrix-variate Gaussian process.}
\label{fig:ranking-gp}
\end{figure}
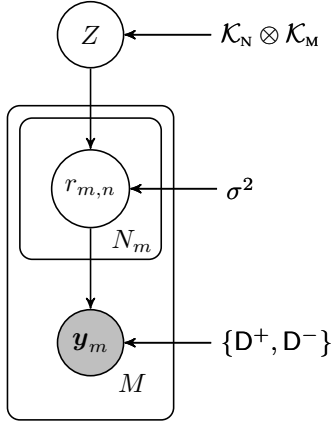
We can now combine the bipartite ranking model with the latent regression model.
The full generative model proceeds as follows (see \figref{fig:ranking-gp}):
\begin{enumerate}
  \item Draw the latent variable $Z$ from a zero mean MV-GP as $Z
  \sim \gp{ {0} }{\cknm}$.
  \item Given $z\smn = Z(\mn)$, draw each score vector independently as
  $r\smn \! \sim \g{z\smn}{\sigmad}$.
  \item For each task $m \in \sfM$, draw the observed response vector $p(\vy_m |
  \vr_m)$ as given by \eqref{eq:generative}.
\end{enumerate}

\subsection{Inference and parameter estimation}\label{sec:inference_and_learning} 
We utilize variational inference
to train the underlying multitask regression model and maximum likelihood to
estimate the parameters of the bipartite ranking model. This is equivalent to
the variational approximation $q(\vr, Z) = \indicator{\vr = \vr^*} q(Z)$ where
$\vr = \{\vr_m\}$.
The variational lower bound of the log likelihood \eqref{eq:var_bnd} is given
by:
\begin{equation*}%
\ln p(\vy|\data) 
\ge  \ln p(\vy|\vr) + \bbE_{\bZ}[\ln p(\vr, \bZ)] - \bbE_{\bZ}[\ln p(\bZ)]
\end{equation*}
where $\vy = \{\vy_m\}$. As outlined in \secref{sec:tracegp}, we restrict our
search to the space for $q(Z)$ of Gaussian processes $q(Z)=\gp{\psi}{S}$
subject to a trace norm constraint $\normht{\psi}{\cK} \le C$. Evaluating
expectations (and ignoring constant terms independent of $\{\vr, \psi, S\}$)
results in the following optimization problem:
\begin{gather}
 \underset{\vr, \bpsi, \bS}{\min} \;
 -\sum_{m \in \sfM} \sum_{l \in \dones_m} \sum_{l' \in \dzero_m} \ln 
 \left( \indicator{r_{m, l} \ge r_{m, l'}} \right) \notag \\
  +\fracl{2\sigmad} \sum_{\mn \in
 \sfT}(r\smn-\psi\smn)^2 +\fracl{2\sigmad} \tr( \bS_{T}) \notag \\
 +\half \trans{\bpsi}\inv{\bK}\bpsi + \half
 \tr(\inv{\bK}\bS) - \ln|\bS| \notag\\
 \quad \quad \text{s.t.}\; \normht{\psi}{\cK} \le C 
 \label{eq:all_cost}
\end{gather}

Inference and parameter estimation follow an alternating optimization scheme. We
alternately optimize each of the parameters $\{\vr, \psi, S\}$ till a local
optima is reached. Following \secref{sec:tracegp}, it is straightforward to show
that the optimal $S$ is given in closed form \eqref{eq:qsigma} and is independent
of $\{\vr, \psi\}$. Hence, model training requires alternating between
optimizing $\vr^* | \psi$ and optimizing $\psi^* | \vr$. We will show that
\eqref{eq:all_cost} is convex, and the alternating optimization approach achieves
the global optimum.
The optimization for $\psi^* | \vr$ follows directly from the discussion in
\secref{sec:tracegp}. Hence, we will focus our efforts on the optimization of
$\vr^* | \psi$.

Collecting the terms of \eqref{eq:all_cost} that are dependent on $\vr$
results in the following loss function for $\vr | \psi$:
\begin{multline*}
 \underset{\vr}{\min} \;
 \overbrace{-\sum_{m \in \sfM}\sum_{l \in \dones_m} 
 \sum_{l' \in \dzero_m} \ln \left( \indicator{r_{m, l} \ge r_{m, l'}} \right) 
 }^\text{Order violation penalty}\\
 + \underbrace{\fracl{2\sigmad} \sum_{\mn \in \sfT}
 (r\smn-\psi\smn)^2}_\text{Square loss}
\end{multline*}
The first term in the loss penalizes violations of order. In fact,
the first term evaluates to infinity if any of the binary order constraints are
violated. Hence, to maximize the log likelihood, the variables $\vr$ must
satisfy the order constraints $\{ \compat{\vr_m}{\sy_m} \; \forall m \in
\sfM\}$. This interpretation suggests a constrained optimization approach:
\begin{equation}\label{eq:min_compat}
  \underset{\{\vr_m| \compat{\vr_m}{\sy_m} \}}{\min} \;
 \half \sum_{l \in \sfT_m}(r_{m, l}-\psi_{m,l})^2 \quad \forall m \in \sfM
\end{equation}
Note that this loss decomposes task-wise. Hence, the proposed approach results in
a list-wise ranking model. We also note that the independence between
tasks means that the optimization is embarrassingly parallel. 

The constrained score vectors $\{\vr_m| \compat{\vr_m}{\sy_m}\}$ can be
optimized efficiently using the inner representation outlined in
\propref{prep:represent}. One issue that arises is that the cost function
\eqref{eq:all_cost} is not invariant to scale. Hence, the loss can be reduced
just by scaling its arguments down. To avoid this degeneracy, we must constrain
the score vectors away from $\mathbf{0}$. We achieve this by constraining the
score vectors to the ordered simplex $\mDo{T_m}$, as it is a convex set and
satisfies the requirement $\mathbf{0} \notin \mDo{T_m}$.
Applying Lemma~\ref{lem:order_simplex}, the score is given by $\vr_m =
\gamma(\sr_m) = \gamma(\bC\vx_m)$ for $\vx_m \in \Do{T_m}$.

Let $\bpsi_m \in \bbR^{T_m}$ be the score vector ordered to satisfy $\bpsi_m  =
[ \{\psi(m,l) \, \forall \,l \in \dones_m\} |\{\psi(m,l') \, \forall \,l' \in
\dzero_m\}] $. The ordering of the score vector is not unique. The loss function
can now be written as:
\begin{equation}\label{eq:min_represent}
 \underset{\vx \in \Do{T_m}}{\min} \;
 \underset{\pi \in \Gamma_m}{\min} \;
 \half \big\| \pi(\bC\vx) - \bpsi_m \big\|_2^2 
 \quad \forall\, m \in \sfM
\end{equation}
This is exactly equivalent to:
\begin{equation}\label{eq:min_representb}
 \underset{\vx \in \Do{T_m}}{\min} \;
 \underset{\gamma \in \Gamma_m}{\min} \;
 \half \big\| \bC\vx - \gamma(\bpsi_m) \big\|_2^2 
 \quad \forall\, m \in \sfM
\end{equation}
The equivalence can be shown simply by setting $\gamma(\cdot) =
\inv{\pi}(\cdot)$. We present both forms as it provides some flexibility when
implementing the algorithm. We optimize \eqref{eq:min_representb} by alternating
optimization. We first optimize the vector $\vx_m^*$ and then optimize
the permutation order $\gamma_m^*$. The overall optimization combining the
variational inference and maximum likelihood parameter estimation is presented
in \aref{alg:als}. The probability vector $\vx_m$ can be optimized efficiently
using the exponentiated gradient (EG) algorithm \cite{kivinen95} or other 
simplex-constrained least squares solvers and can be embarrassingly parallelized 
over the tasks $\sfT_m$. Optimization of $\gamma_m$ requires optimizing
over all permutations of the vector. This may be na\"{i}vely solved by expensive
enumeration or by solving a combinatorial assignment problem.
\begin{lemma}[Optimality of sorting \cite{acharyya12}]\label{lem:sort}
If $x_1 \geq x_2$ and $y_1 \geq y_2$, then
$\normbs{ \prta{x} - \prta{y}} \leq \normbs{ \prta{x} - \prtb{y} }$ and
$\normbs{ \prta{y} - \prta{x}} \leq \normbs{ \prtb{y} - \prta{x} }$.
\end{lemma}
Lemma~\ref{lem:sort} implies that selecting the pair-wise sorted order for
equivalent items minimizes the loss. Lemma~\ref{lem:sort} can
then be extended to $\vy \in \bbR^d$ using induction over $d$. Hence, selecting
$\gamma_m$ as the sorted ordering in each block of equivalent values $\{
\dones_m, \dzero_m\}$ minimizes the loss.
\begin{Algorithm}
\begin{algorithmic}[1]
\STATE \textbf{initialize} $\psi$, $\{\vx_m\}$, $\{\gamma_m\}$
	\REPEAT
	\STATE Update $\psi^*| \vr$ by solving \eqref{eq:qmean_reg}.
	\FORALL{$m \in \sfM$}
		\STATE Update $\gamma_m^* | \bpsi_m$ by block sorting (\lemref{lem:sort}).
		\STATE Update $\vx_m^* | \gamma_m, \bpsi_m$ by
		solving \eqref{eq:min_representb}.
	\ENDFOR
	\UNTIL{converged}
\RETURN $\psi$, $\{\vx_m\}$, $\{\gamma_m\}$
\STATE Compute $S$ using \eqref{eq:qsigma}
\end{algorithmic}
\caption{Variational inference and maximum likelihood parameter
estimation.}\label{alg:als}
\end{Algorithm}

The set of compatible vectors as defined by \propref{prep:represent} is a convex
cone, i.e., the convex composition $\vz = \alpha \vu + (1-\alpha) \vv, \, \alpha
\in [0, 1] $ of vectors in the set remains in the set, and any scaling $\alpha
\vz$ where $\alpha >0$ preserves compatibility. To show convexity of parameter
estimation, it remains to show that the combination of optimizing $\vx_m$ and
optimizing $\gamma_m$ minimizes \eqref{eq:min_compat}.
This is shown using the following lemma.

\begin{lemma}[Convexity of parameter estimation
\cite{acharyya12}]\label{lem:r_convex} 
Let $\vr$ be partitioned into
two sets such that $\vr_1 = \{r_k,\; \forall k\in \dones\}$ and $\vr_2 =
\{ r_k,\; \forall k \in \dzero\}$, and let:
\begin{equation*}
\prta{\vr}^* = \underset{\substack{\vr'_i \in \Pi(\vr_i)\\
\vr'_1 \geq \vr'_2}}{\arg \min} \normbs{\prtnb{\vr}{1}{2}{'} - \prta{\vz} },
\end{equation*}
where $\Pi(\vr_i)$ is the set of all permutations of the vector $\vr_i$ and
$\vr'_1 \geq \vr'_2$ represents element wise inequality, then $\vr_i^*$ is
isotonic with $\vz_i \; \forall i=1,2$.
\end{lemma}

We can now consider the global properties of the variational inference and
maximum likelihood parameter estimation.
\begin{lemma}[Joint convexity]\label{lem:all_convex} 
The variational inference and parameter estimation given by
\eqref{eq:all_cost} is jointly convex in $\{ \psi, S, \vr \}$. Alternating
optimization (\aref{alg:als}) recovers the global optimum.
\end{lemma}
{\bf Sketch of proof:}
Recall that squared loss is jointly convex in both of its arguments. In
addition, the components $\{\psi, S\}$ and $ \vr$ are in separate convex
sets. Hence to show global convexity, it is sufficient to show that
\eqref{eq:all_cost} is convex separately in $\{\psi, S\}$ (by \thref{thrm:gp})
and $\vr$ (by \lemref{lem:r_convex}). It follows from the joint convexity of
\eqref{eq:all_cost} that the alternating optimization of \aref{alg:als}
recovers the global optimum.

The proposed model is trained to estimate bipartite ranking scores for each task
and the underlying multitask latent regression distribution. Item rankings
are predicted by sorting the expected noise-free scores of the trained model
$\bbE[z\smn| \data] = \psi(m, n)$.

\section{Experiments}\label{sec:experiments}
This section details the experiments comparing the performance of the
proposed model applied to the disease-gene prioritization task. We evaluated the
modeling performance on association data curated from the OMIM database
\cite{mckusick07} by the authors of \cite{mordelet11} and data we
curated ourselves. We partitioned each dataset into five-fold cross validation
sets. The model was trained on $4$ of the $5$ sets and tested on the held out
set. The results presented are the averaged $5$-fold cross validation
performance. Great care was taken to train all the models on the same datasets.
Hence the results represent performance differences due to either the low rank
modeling, the list-wise bipartite ranking model, or both.

{\bf Baseline (ProDiGe \cite{mordelet11}):} We compared our proposed model to
ProDiGe which, to the best of our knowledge, is the state of the art in the
disease-gene prioritization literature. ProDiGe estimates the prioritization
function using multitask support vector machines trained with gene kernel and
disease kernel information. Parameter selection for ProDiGe was performed as
suggested by the authors \cite{mordelet11}.

{\bf OMIM dataset:} The OMIM dataset \cite{mckusick07} is a curated database of
known human disease-gene associations ($4178$ associations in the provided
dataset). We derived the gene-gene interaction graph using data from HumanNet
\cite{lee11}. We selected all genes with one or more connections in the network
and all diseases with one or more genetic associations. This resulted in a
disease-gene matrix with $M = 3,210$ diseases, $N=13,614$ genes and $T=3,636$
known associations (data sparsity $.0083\%$). In addition, the gene-gene graph
contained $433,224$ known gene-gene links.
We note the extreme sparsity of this matrix, and the resulting difficulty of the
ranking task. Such sparse datasets are typical in the disease-gene domain. The
OMIM dataset did not contain a disease graph; hence, we were unable to test the
generalization of the methods to new diseases.

{\bf Curated dataset:} We curated a large disease-gene association dataset. The
set of genes were defined using the NCBI ENTREZ Gene database \cite{maglott11},
and the set of diseases were defined using the ``Disease'' branch of the NIH
Medical Subject Heading (MeSH) ontology \cite{mesh}. We extracted co-citations
of these genes and diseases in the PubMed/Medline database \cite{pubmed} to
identify positive instances of disease-gene associations. We derived our
gene-gene interaction graph using data from HumanNet \cite{lee11} and our
disease-disease similarity graph from the MeSH ontology. This resulted in a set
of $250,190$ observed interactions, $21,243$ genes and $4,496$ diseases. We
selected all genes with one or more connections in the gene-gene graph and all
diseases with with one or more connections in the disease-disease graph. This
resulted in a dataset with $M = 4,495$ diseases, $N=13,614$ genes and
$T=224,091$ known associations (data sparsity $0.36\%$). The resulting disease network
contained $13,922$ links, and the gene network contained $433,224$ links.

We were unable to run ProDiGe on the full dataset due to insufficient memory for
storing the kernel matrix. Instead, we trained ProDiGe and the MV-GP models on a
randomly selected 5\% subsample of the associations. 
We also provide results for the MV-GP models trained on the full dataset.
We performed two kinds of experiments for the curated dataset. The first
experiment ({\bf known diseases}) tests the ranking ability of the model for
associations selected randomly over the matrix. The second experiment ({\bf new
diseases}) tests the generalization ability of the model for new diseases not
observed in the training set. For the known disease experiments, the cross
validation associations were randomly selected over the matrix. For the new
disease experiments, the cross validation was performed row-wise, i.e., we
selected training set diseases and test set diseases.

{\bf Model Setup:} The proposed model was trained using the alternating
optimization approach (\aref{alg:als}). The trace constrained mean function was
estimated using the cost function \eqref{eq:qmean_reparam}. The model was
trained using our implementation of the algorithm outlined in \cite{laue12}.
Like other large scale trace constrained matrix optimizers, \cite{laue12}
maintains a low rank representation. The rank is estimated automatically by the
optimizer. We found that employing a row bias improved the model performance, so
we learned row biases while training. Note that the row offsets do not change
the ranking and hence are not required for testing.

We selected the hyperparameter $\lambda = s*\lambda_{\max}$ with $30$ values of
$s$ logarithmically spaced between $10^{-3}$ and $1.0$.  Let $F(\bB)$ be the
loss function. Then $\lambda_{\max}$ is the maximum singular value of
$\D{F(\bB)}{\bB}\big|_{\bB = \mathbf{0} }$. The optimization returns the zero
matrix for any $\lambda > \lambda_{\max}$. \cite{dudik12}. We used warm start to
speed up the computation for decreasing values of $s$. We selected $\alpha \in
\{1,\, 0.8,\, 0.6,\, 0.4,\, 0\}$. 

We implemented the full rank Gaussian process model ($\alpha=0$) by keeping the
kernels as separate row and column kernels. This allowed us to scale the model
to the larger datasets at the expense of more computations.
We observed that the full rank model required a significant amount of
computation time. This observation provides further motivation for the low rank
approach. The full rank Gaussian process  was trained directly using the
Broyden-Fletcher-Goldfarb-Shanno (BFGS) algorithm.

{\bf Sampling unknown negative items:} Recall that the observed data only
consists of known associations. Following ProDiGe \cite{mordelet11} we sampled
``negative'' observations randomly over the disease-gene association matrix.
We sampled $10$ different negatively labeled item sets. All models are trained
with the positive set combined with one of the negative labeled sets.
The model scores are computed by averaging the scores over the 10 trained
models. All algorithms were trained using the same samples.

{\bf Covariance/Kernels:} The covariances for the MV-GP prior and the kernels
for ProDiGe were computed from gene graph $\cG\sm$ and the disease graphs
$\cG\sn$.
We performed preliminary experiments with a large class of graph kernels
\cite{smola03} and selected the exponential kernel.
We briefly outline kernel generation for the gene kernel. Let $\bA\sm$ be the
adjacency matrix for the gene-gene graph. We computed the normalized Laplacian matrix
as $\bL\sm = \bI - \bD^{-\half}\bA\sm\bD^{-\half}$, where $\bI$ is the identity
matrix and $\bD$ is a diagonal matrix with entries $\bD_{i,i}=
(\bA\sm\mathbf{1})_i$. The exponential kernel is given by $\bK'\sm =
\exp{(-\bL\sm)}$. Following the suggestion in \cite{mordelet11} and preliminary
experiments, we observed an improvement in performance with the identity matrix
added to the exponential kernel. Hence the final kernel is given by $\bK\sm =
\exp{(-\bL\sm)} + \bI$. The disease kernel generation was obtained using the
same approach when a disease-disease graph was available. All algorithms were trained
using the same kernel matrices.

{\bf Metrics:} Experimental validation of disease-gene associations in a
laboratory can be time consuming and costly, so only a small set of the top
ranked predictions are of practical interest. Hence, we focus on metrics that
capture the ranking behavior of the model at the top of the ranked list. In
addition, all metrics are computed on the test set after removing all relevant
genes that had been observed in the training set removed. All metrics are
computed per disease and then averaged over all the diseases in the test set.
Let $\sg_l$ denote the labels of item (gene) $l$ as sorted by the predicted
scores of the trained regression model, and let $G_m = |\dones_m| = \sum_{l}
\indicator{\sg_l=1}$ be the total number of relevant genes for disease $m$ in
the test data  after removing relevant genes observed in the training data.
The metrics computed are as follows:
\begin{enumerate}
  \item Area under the ROC curve (AUC) \eqref{eq:auc}. This measures the overall
  ranking performance of the model.
  \item The precision at $k\in\{1, 2, \ldots , 100\}$ computes the fraction of
  relevant genes retrieved out off all retrieved genes at position
  $k$. 
  \begin{equation*} 
  	\pres{k} = \frac{\sum_{l=1}^k \indicator{\sg_l=1} }{k}.
  \end{equation*}
  \item The recall at $k\in\{1, 2, \ldots , 100\}$ computes the fraction of
  relevant genes retrieved out off all relevant genes.
  \begin{equation*} 
  	\recall{k} = \frac{\sum_{l=1}^k \indicator{\sg_l=1} }  
  	{\min(G_m, k) }.
  \end{equation*}
  \item Mean average precision at $k=100$ ($\map{100}$) computed as the mean of
  the average precision at $k=100$. The average precision is given as:
  \begin{equation*} 
  \ap{k} = \frac{\sum_{l=1}^{k} \indicator{\sg_l=1} \pres{l} }
  {\min(G_m, k) }
  \end{equation*}
\end{enumerate}
The $\map_{@100}$ metric was used for model selection over the cross
validation runs. To reduce notation, $\map$ refers to $\map_{@100}$ in all
results. Higher values reflect better performance for the $\auc,\, \pres{k},\,
\recall{k}$ and $\map$ metrics, and their maximum value is $1.0$.
\subsection{Discussion}
We present performance results for ProDiGe, the standard MV-GP (\emph{Hilbert},
$\alpha=0$), the trace norm regularized MV-GP (\emph{Trace}, $\alpha=1$), and
the best overall MV-GP model (\emph{Best}). 

\begin{figure}[t]
\begin{center}
\includegraphics[width=0.8\columnwidth]{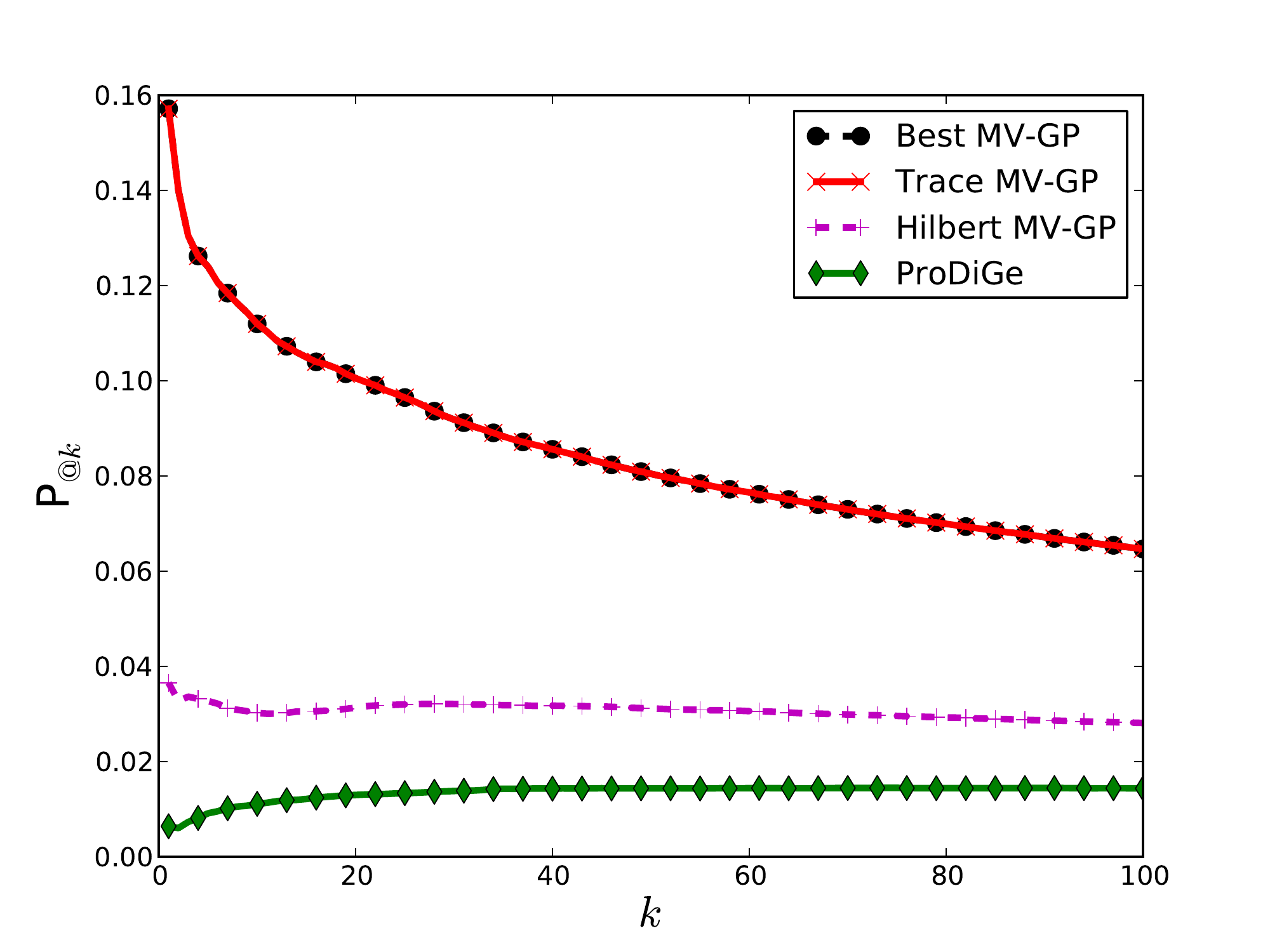}
\includegraphics[width=0.8\columnwidth]{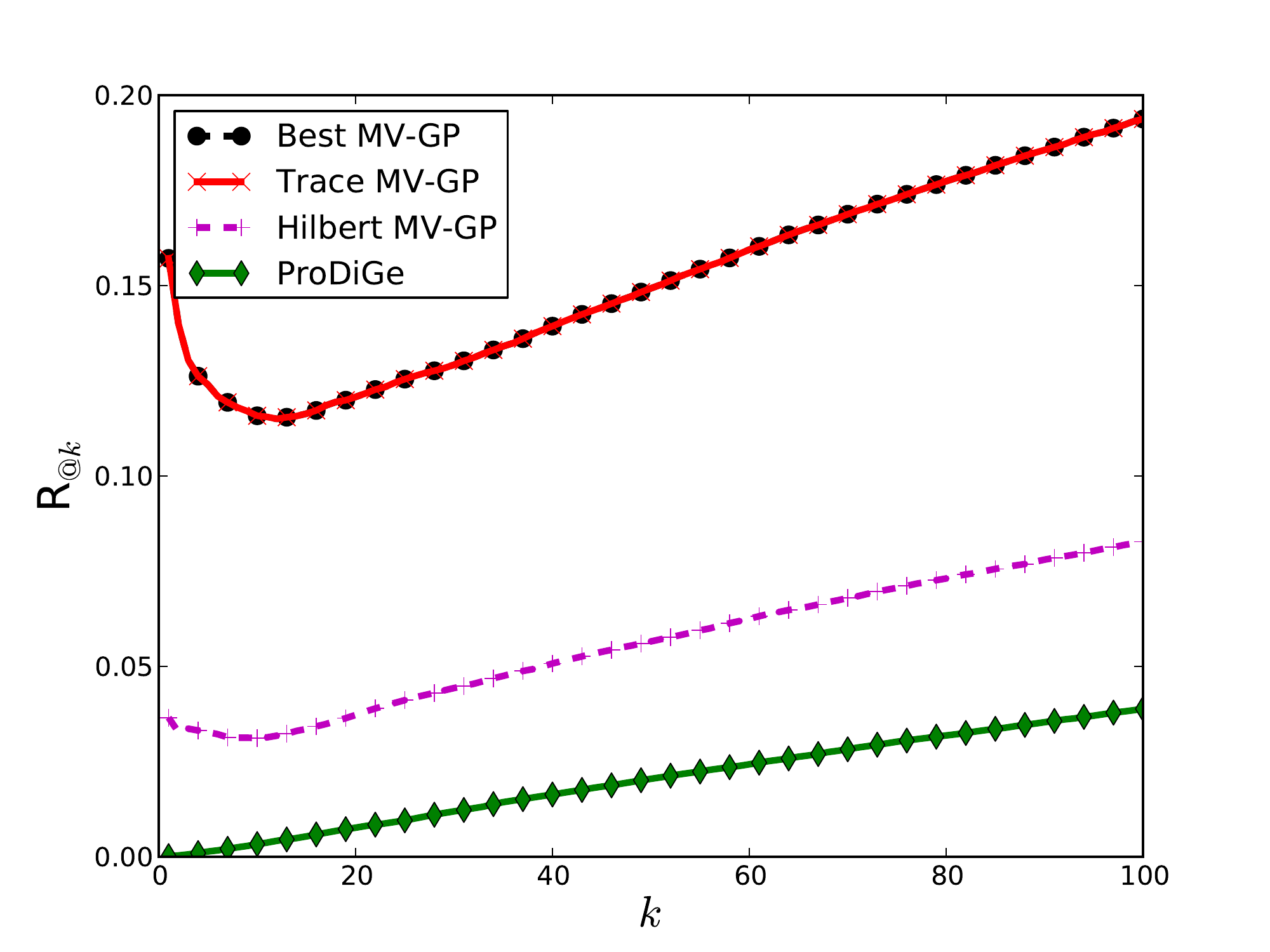}
\end{center}
\caption{Curated data (known diseases, 5\% subsample)  experiment results:
 precision (top) and recall (bottom) curves $@k=1, 2, \ldots, 100$. Best and
 Trace curves overlap.}
\label{fig:our_mat_p5_prec_rec}
\end{figure}

\begin{figure}[t]
\begin{center}
\includegraphics[width=0.8\columnwidth]{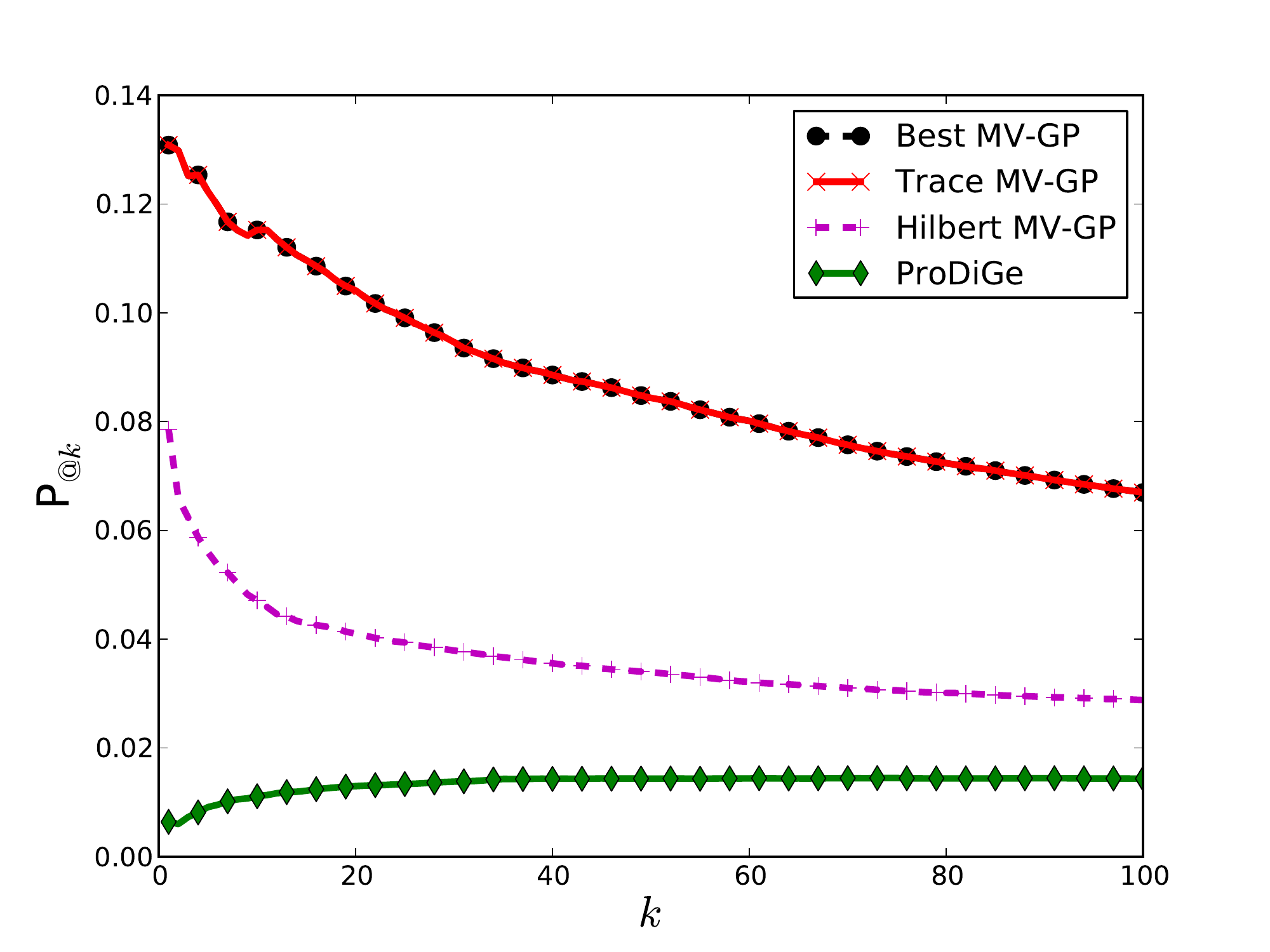}
\includegraphics[width=0.8\columnwidth]{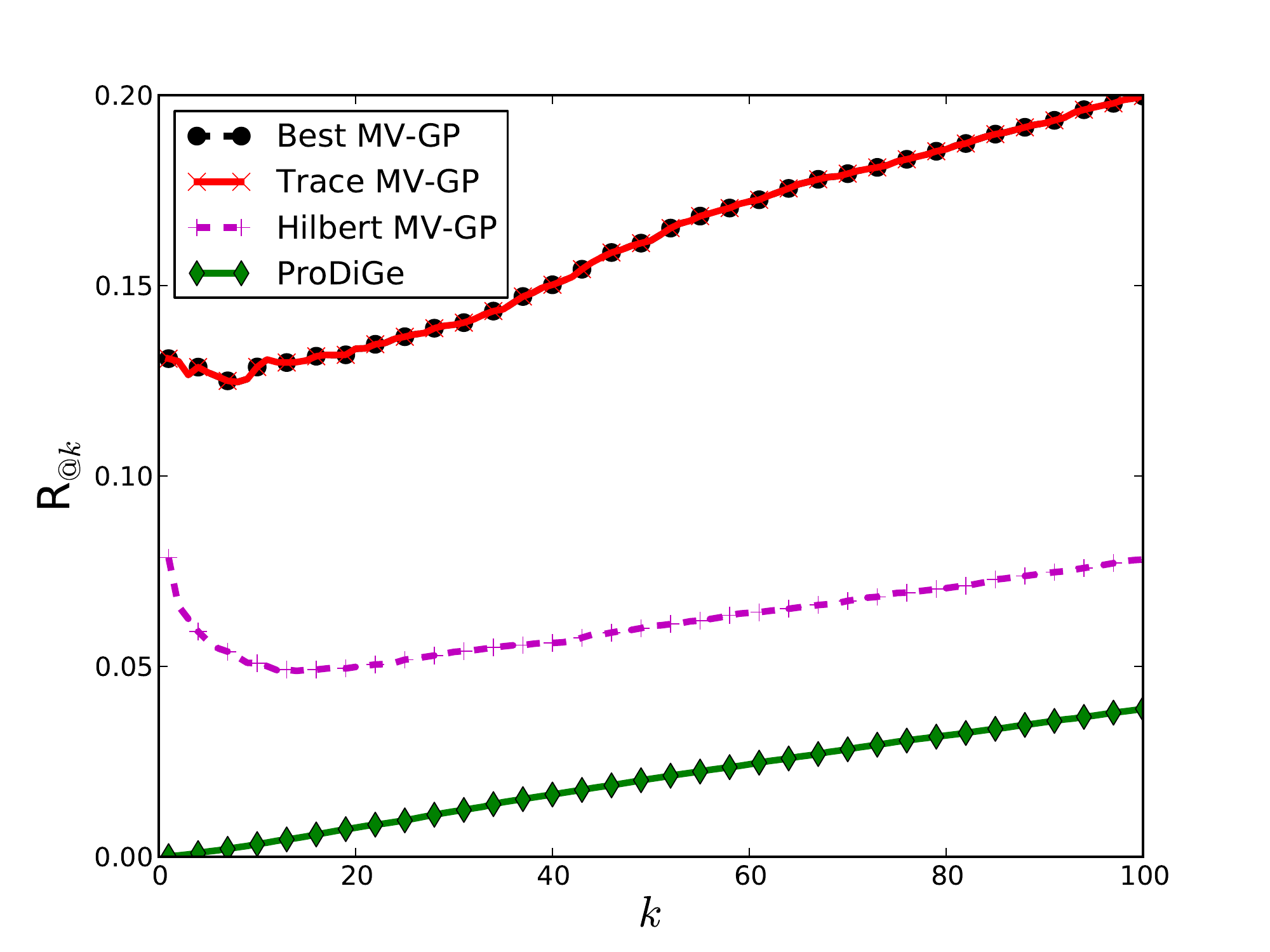}
\end{center}
\caption{Curated data (new diseases, 5\% subsample)  experiment results:
 precision (top) and recall (bottom) curves $@k=1, 2, \ldots, 100$. Best and
 Trace curves overlap. }
\label{fig:our_cold_p5_prec_rec}
\end{figure}

Our first experiment was on the known disease prediction with the 5\% subset of
the curated data. This task is very challenging as the training data consisted
of an average of less than $3$ known associations out the possible $13,614$ per
disease. The difficulty of this task is reflected in the performance results
shown in \tabref{tab:our_5_mat} and \figref{fig:our_mat_p5_prec_rec}. We found
that the trace model had the same performance as the best MV-GP model, suggesting
that the trace norm is an effective regularization in this case. We found that
the trace regularization resulted in a significant improvement in performance
across metrics compared to ProDiGe and the Hilbert models. We also experimented
with predicting the gene ranking of new diseases not seen during training and
found similar performance as shown in \tabref{tab:our_5_cold} and
\figref{fig:our_cold_p5_prec_rec}. As this is new disease prediction, none of
the known genes on are removed from the test diseases. Interestingly, we found
that this seems to improve the model performance as compared to the in-matrix
prediction.

\begin{figure}[t]
\begin{center}
\includegraphics[width=0.8\columnwidth]{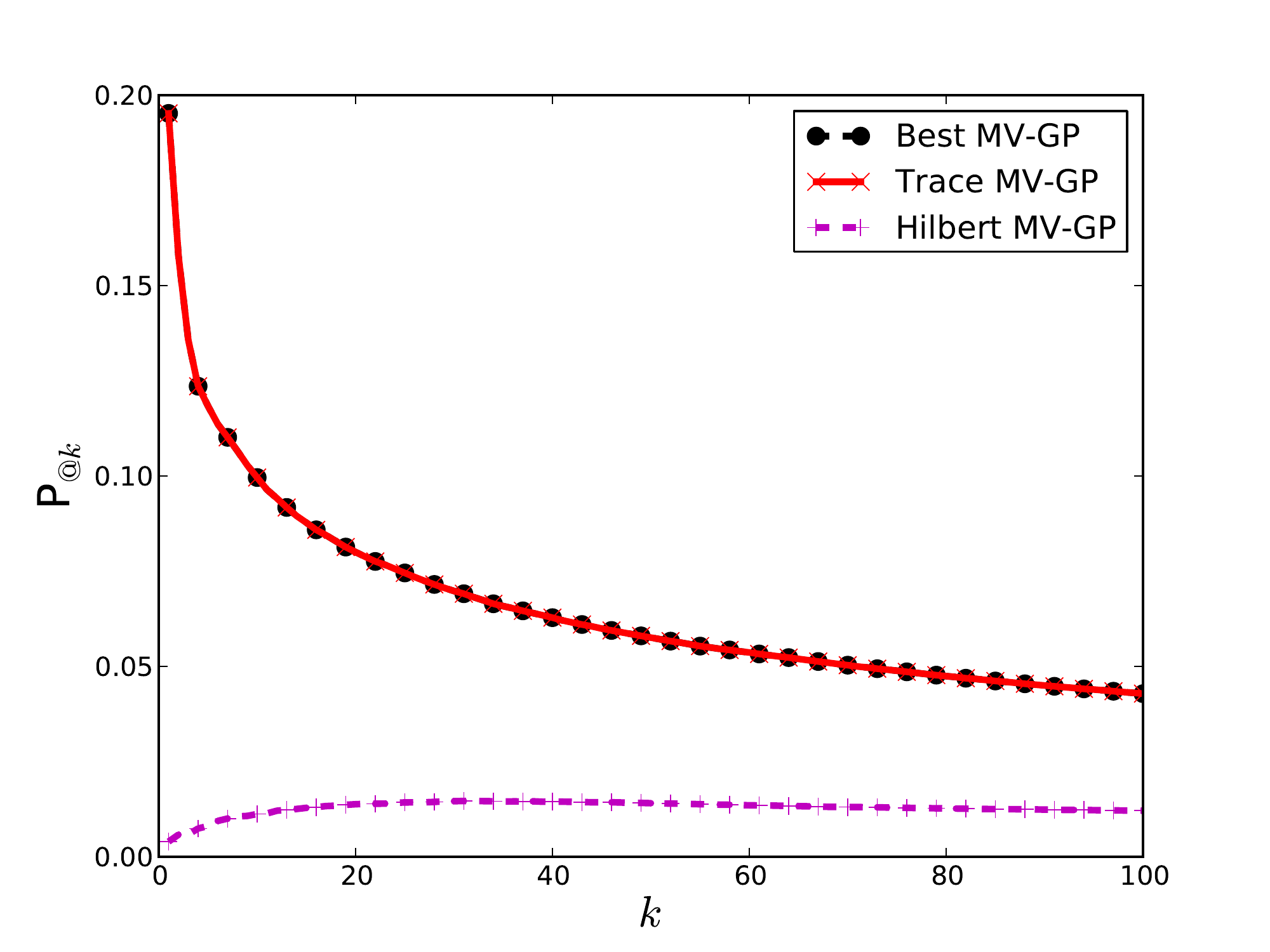}
\includegraphics[width=0.8\columnwidth]{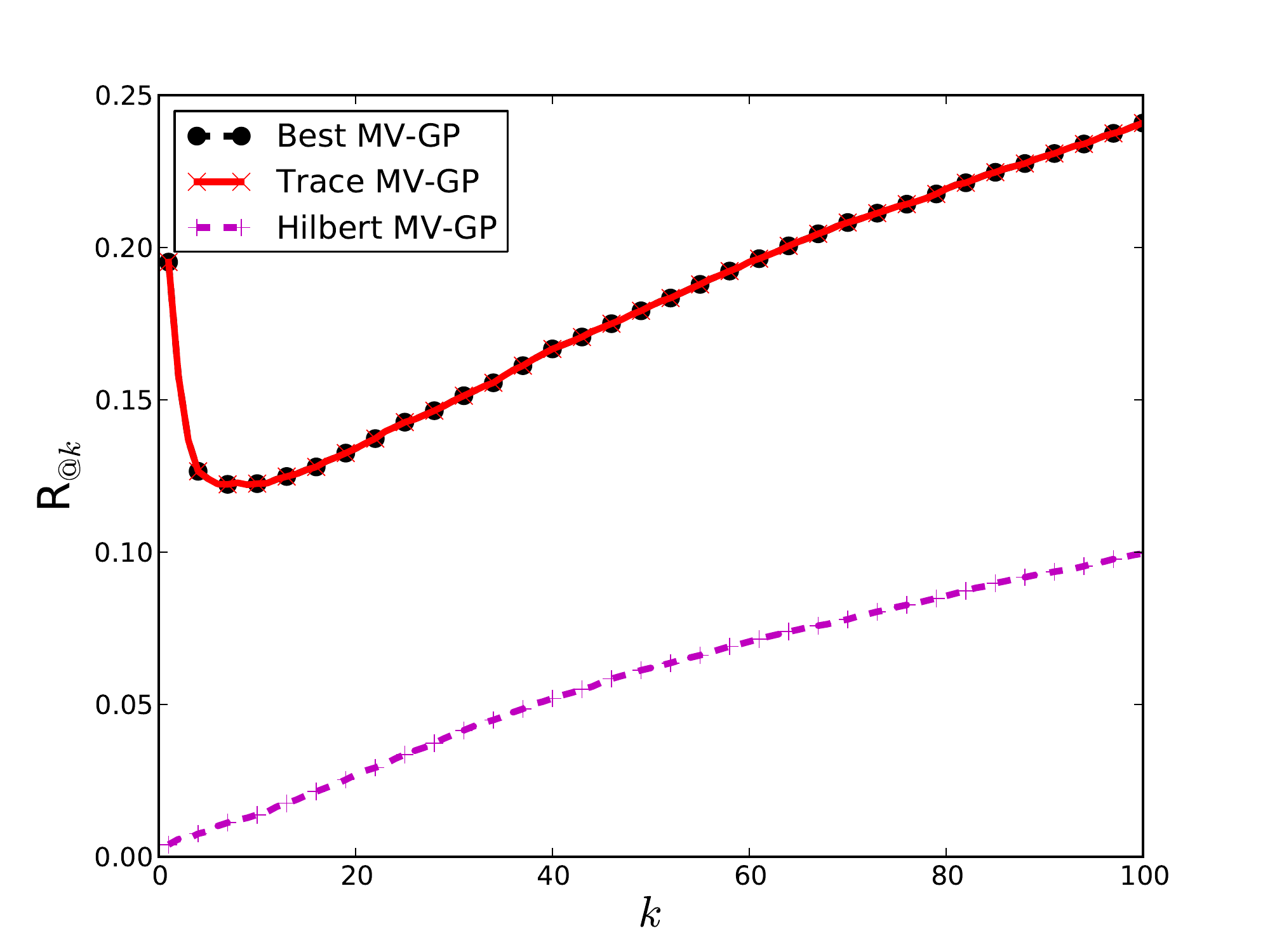}
\end{center}
\caption{Curated data (known diseases, full dataset)  experiment results:
 precision (top) and recall (bottom) curves $@k=1, 2, \ldots, 100$. Best and
 Trace curves overlap. }
\label{fig:our_mat_prec_rec}
\end{figure}

\begin{figure}[t]
\begin{center}
\includegraphics[width=0.8\columnwidth]{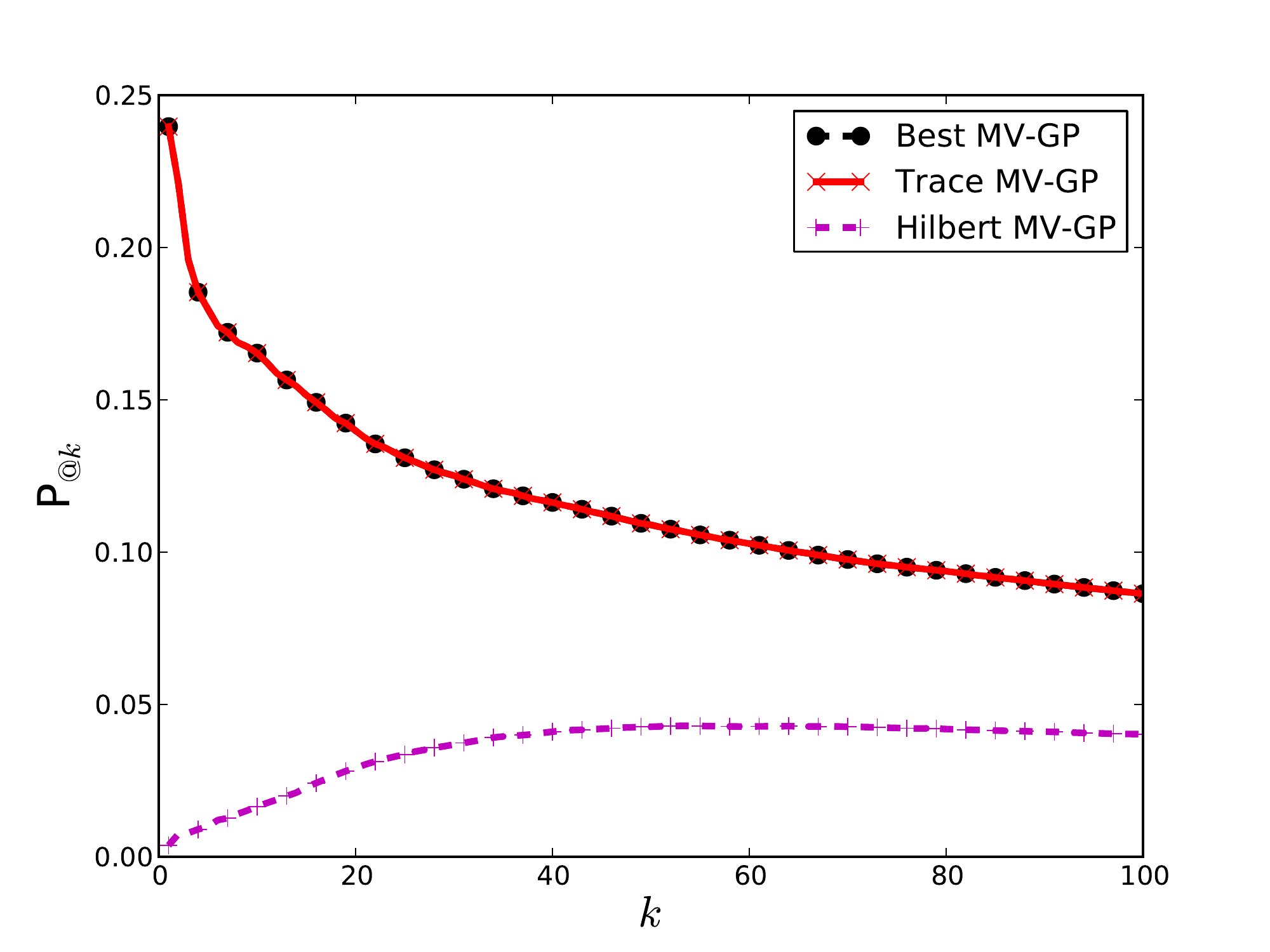}
\includegraphics[width=0.8\columnwidth]{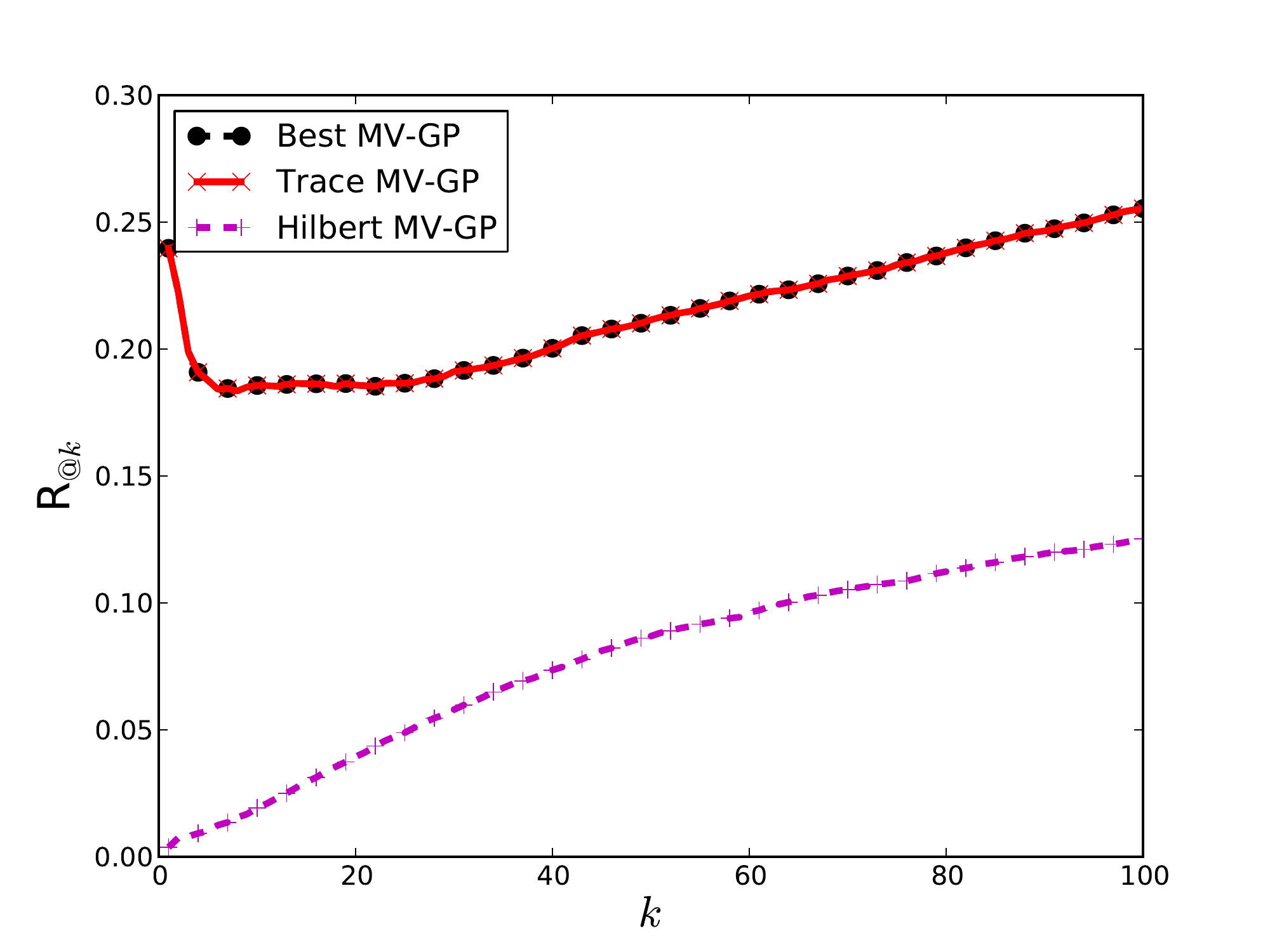}
\end{center}
\caption{Curated data (new diseases, full dataset)  experiment results:
 precision (top) and recall (bottom) curves $@k=1, 2, \ldots, 100$. Best and
 Trace curves overlap.}
\label{fig:our_cold_prec_rec}
\end{figure}

Next, we experimented with prediction on the full curated dataset predicting
known diseases. We were unable run this experiment with ProDiGe due to memory
limitations. Hence, only results from the proposed model are shown. The results are as
shown in \tabref{tab:our_mat} and \figref{fig:our_mat_prec_rec}.
As expected, we observed a significant improvement in performance by using the
entire dataset. Similar results were observed for the new disease prediction as
shown in \tabref{tab:our_cold} and \figref{fig:our_cold_prec_rec}. In all
models, we observed that the trace norm constrained approach out-performed the
standard full rank MV-GP model. We especially note the performance improvement
at the top of the list, as these are the most important to the domain.

\begin{figure}[t]
\begin{center}
\includegraphics[width=0.8\columnwidth]{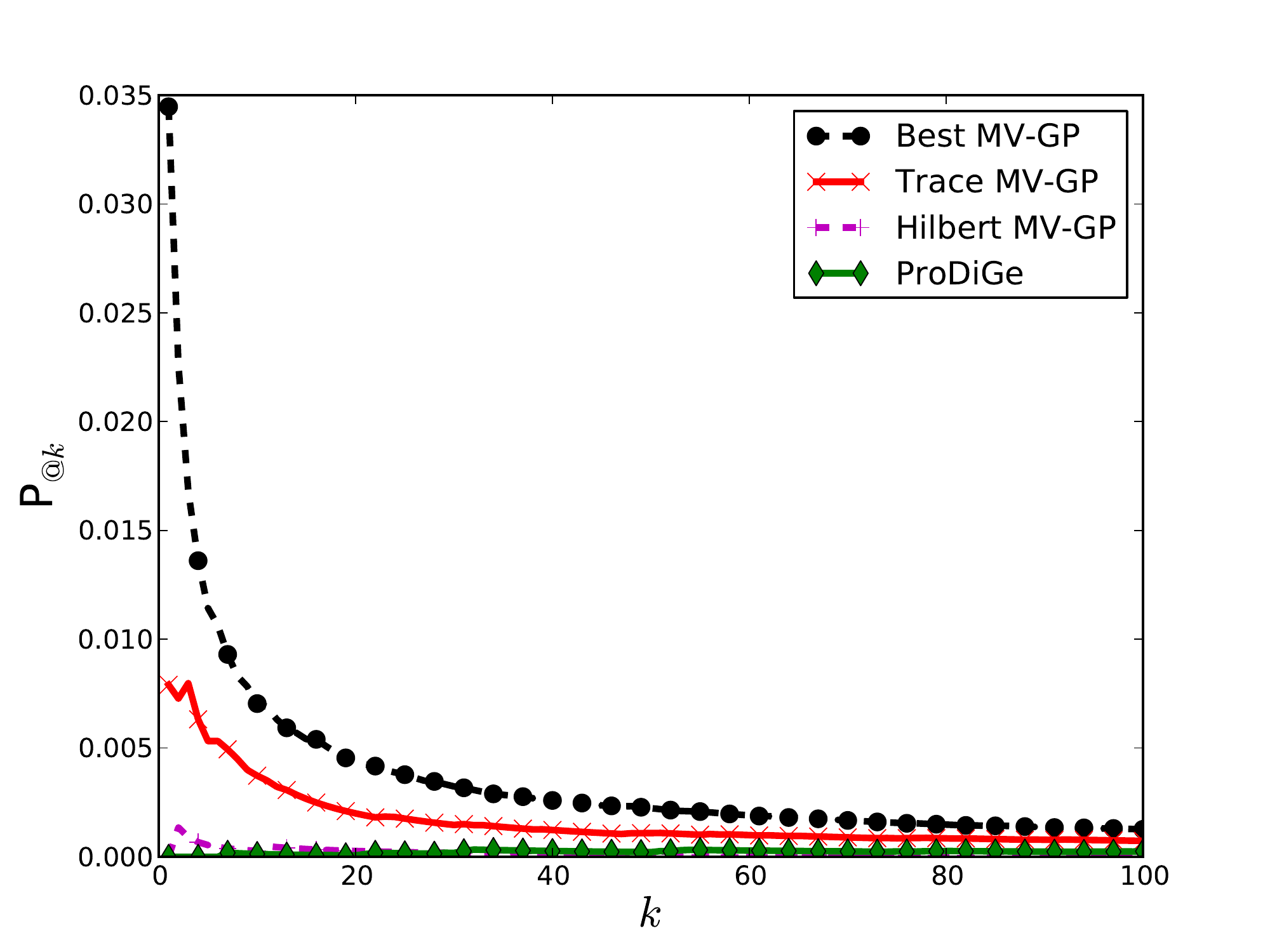}
\includegraphics[width=0.8\columnwidth]{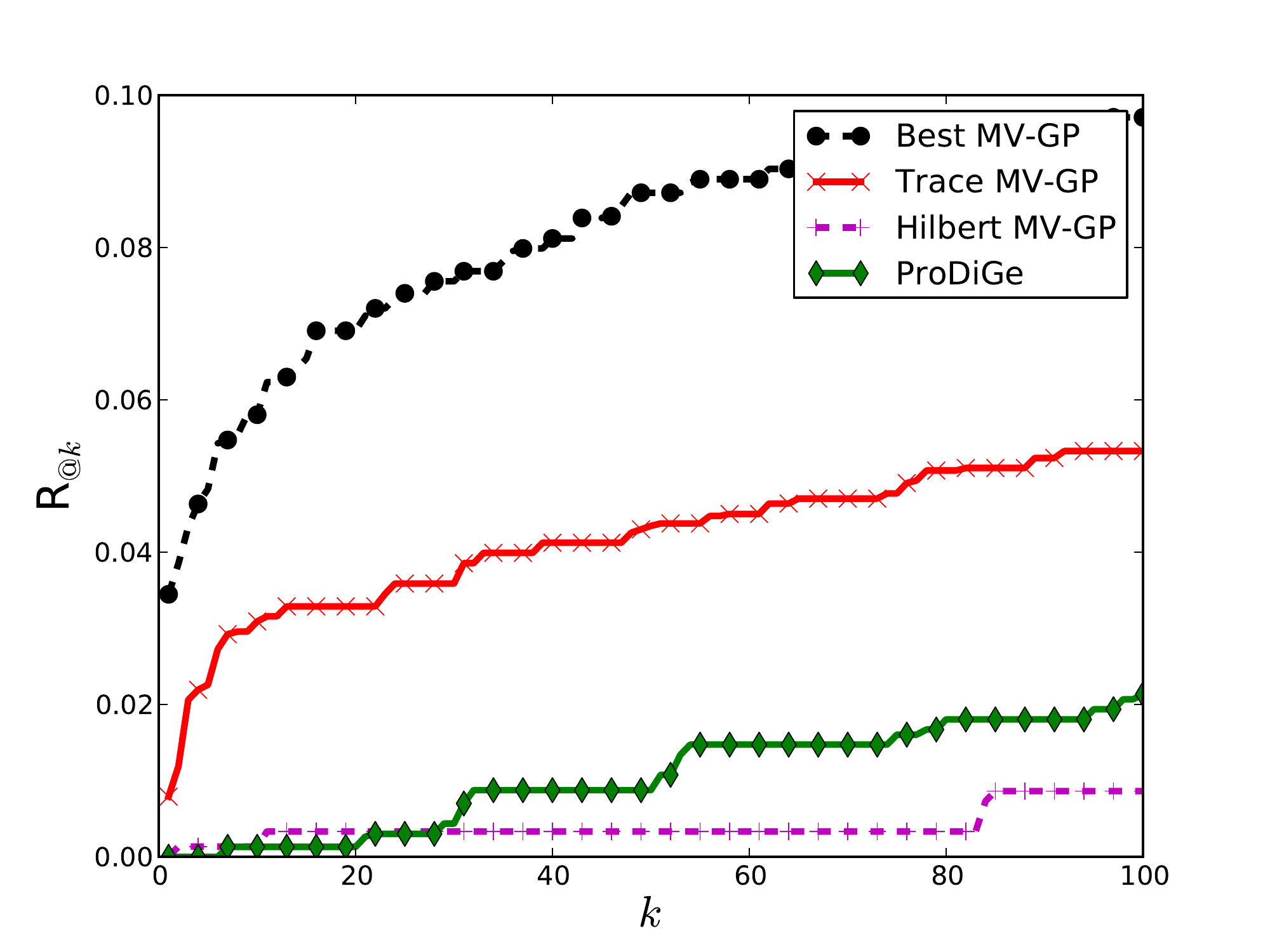}
\end{center}
\caption{OMIM data experiment results: precision (top) and recall (bottom)
curves $@k=1, 2, \ldots, 100$.}
\label{fig:omim_prec_rec}
\end{figure}

Our final experiment was on the OMIM dataset. The results are as shown in
\tabref{tab:omim} and \figref{fig:omim_prec_rec}. These results are especially
interesting as we found that the best overall model outperformed the trace
model, and significantly outperformed all other model in terms of ranking at the
top of the list. This suggests that the spectral elastic net regularizer may be
most useful with significant data sparsity. ProDiGe out-performed the Hilbert
model in terms of recall, but  Hilbert model had the best overall ranking
performance as measured by AUC. We are investigating this observation further,
but preliminary investigation suggests that the metrics are more sensitive to
small changes in order when the data is very sparse. The sparsity also explains
the significant drop in $\pres{k}$ as k grows.


\begin{table}[!t]
\renewcommand{\arraystretch}{1.3}
\caption{Curated data experiment (known diseases, 5\% subsample) 
avg. (std.) performance comparison.}
\label{tab:our_5_mat}
\begin{center}
\begin{tabular}{|l||c|c|c|c|}
\hline
&Best & Trace & Hilbert & ProDiGe \\
\hline\hline
\!\!\auc\!\!\! & {\bf 0.793 (0.002)} & {\bf 0.793 (0.002)} & {0.687 (0.002)} &
{0.716 (0.001)} \\
\hline
\!\!\map\!\!\! & {\bf 0.042 (0.003)} & {\bf 0.042 (0.003)} & {0.009 (0.001)} &
{0.003 (0.000)} \\
\hline
\!\!$\pres{100}$\!\!\! & {\bf 0.065 (0.001)} & {\bf 0.065 (0.001)} & {0.028
(0.001)} & {0.014 (0.000)} \\
\hline
\!\!$\recall{100}$\!\!\! & {\bf 0.194 (0.001)} & {\bf 0.194 (0.001)} & {0.083
(0.003)} & {0.039 (0.002)} \\
\hline

\end{tabular}
\end{center}
\end{table}
\begin{table}[!t]
\renewcommand{\arraystretch}{1.3}
\caption{Curated data experiment (new diseases, 5\% subsample)
avg. (std.) performance comparison.}
\label{tab:our_5_cold}
\begin{center}
\begin{tabular}{|l||c|c|c|c|}
\hline
&Best & Trace & Hilbert & ProDiGe \\
\hline\hline
\!\!\auc\!\!\! & {\bf 0.822 (0.014)} & {\bf 0.822 (0.014)} & {0.661 (0.018)} &
{0.716 (0.001)} \\
\hline
\!\!\map\!\!\! & {\bf 0.047 (0.009)} & {\bf 0.047 (0.009)} & {0.013 (0.004)} &
{0.003 (0.000)} \\
\hline
\!\!$\pres{100}$\!\!\! & {\bf 0.067 (0.014)} & {\bf 0.067 (0.014)} & {0.029
(0.009)} & {0.014 (0.000)} \\
\hline
\!\!$\recall{100}$\!\!\! & {\bf 0.200 (0.019)} & {\bf 0.200 (0.019)} & {0.078
(0.011)} & {0.039 (0.002)} \\
\hline

\end{tabular}
\end{center}
\end{table}
\begin{table}[!t]
\renewcommand{\arraystretch}{1.3}
\caption{Curated data experiment (known diseases, full dataset) 
avg. (std.) performance comparison.}
\label{tab:our_mat}
\begin{center}
\begin{tabular}{|l||c|c|c|}
\hline
&Best & Trace & Hilbert \\
\hline\hline
\!\!\auc\!\!\! & {\bf 0.869 (0.001)} & {\bf 0.869 (0.001)} & {0.782 (0.001)}
\\
\hline
\!\!\map\!\!\! & {\bf 0.054 (0.001)} & {\bf 0.054 (0.001)} & {0.006 (0.000)}
\\
\hline
\!\!$\pres{100}$\!\!\! & {\bf 0.043 (0.000)} & {\bf 0.043 (0.000)} & {0.012
(0.000)}
\\
\hline
\!\!$\recall{100}$\!\!\! & {\bf 0.241 (0.001)} & {\bf 0.241 (0.001)} & {0.100
(0.001)}
\\
\hline

\end{tabular}
\end{center}
\end{table}
\begin{table}[!t]
\renewcommand{\arraystretch}{1.3}
\caption{Curated data experiment (new diseases, full dataset)
avg. (std.) performance comparison.}
\label{tab:our_cold}
\begin{center}
\begin{tabular}{|l||c|c|c|}
\hline
&Best & Trace & Hilbert \\
\hline\hline
\!\!\auc\!\!\! & {\bf 0.871 (0.009)} & {\bf 0.871 (0.009)} & {0.787 (0.015)}
\\
\hline
\!\!\map\!\!\! & {\bf 0.080 (0.018)} & {\bf 0.080 (0.018)} & {0.013 (0.003)}
\\
\hline
\!\!$\pres{100}$\!\!\! & {\bf 0.086 (0.021)} & {\bf 0.086 (0.021)} & {0.040
(0.010)}
\\
\hline
\!\!$\recall{100}$\!\!\! & {\bf 0.255 (0.021)} & {\bf 0.255 (0.021)} & {0.125
(0.013)}
\\
\hline

\end{tabular}
\end{center}
\end{table}
\begin{table}[!t]
\renewcommand{\arraystretch}{1.3}
\caption{OMIM data experiment avg. (std.) performance comparison.}
\label{tab:omim}
\begin{center}
\begin{tabular}{|l||c|c|c|c|}
\hline
&Best & Trace & Hilbert & ProDiGe \\
\hline\hline
\!\!\auc\!\!\! & {0.654 (0.028)} & {0.649 (0.029)} & {\bf 0.686 (0.016)} &
{0.524 (0.018)} \\
\hline
\!\!\map\!\!\! & {\bf 0.041 (0.008)} & {0.015 (0.002)} & {0.001 (0.001)} &
{0.001 (0.000)} \\
\hline
\!\!$\pres{100}$\!\!\! & {\bf 0.001 (0.000)} & {0.001 (0.000)} & {0.000 (0.000)}
& {0.000 (0.000)} \\
\hline
\!\!$\recall{100}$\!\!\! & {\bf 0.097 (0.014)} & {0.053 (0.018)} & {0.009
(0.003)} & {0.021 (0.005)} \\
\hline

\end{tabular}
\end{center}
\end{table}

\section{Conclusion}\label{sec:conclude}
This paper proposes a novel hierarchical model for multitask bipartite ranking
that combines a trace constrained matrix-variate Gaussian process and a
bipartite ranking model. We showed that the trace constraint led to a mean
function with low rank and discussed the spectral elastic net as the MAP
regularizer that arises from this model. We showed that constrained variational
inference for the Gaussian process combined with maximum likelihood parameter
estimation for the ranking model was jointly convex. We applied the proposed
model to the prioritization of disease-genes and found that the proposed model
significantly improved performance over strong baseline models.

We plan to explore the trace norm constrained MV-GP and the spectral elastic net
further and analyze their theoretical properties. We also plan to explore
parameter estimation using the resulting constrained posterior distribution.
In addition, we plan to investigate the applications of the constrained MV-GP to
other tasks including multitask regression and collaborative filtering.
\ifCLASSOPTIONcompsoc
  \section*{Acknowledgments}
\else
  \section*{Acknowledgment}
\fi 
Authors acknowledge support from NSF grant IIS 1016614. We thank Sreangsu
Acharyya for helpful discussions on bipartite ranking. We also thank U.~Martin
Blom and Edward Marcotte for providing the OMIM data set.

\bibliographystyle{IEEEtran}
\bibliography{IEEEabrv,rank_GP}

\begin{IEEEbiographynophoto}{Oluwasanmi Koyejo}
is a PhD student working on data mining and machine learning at The
University of Texas at Austin and advised by Dr. Joydeep Ghosh. He received his
B.S. Degree in Electrical Engineering with a minor in Statistics from the New
Jersey Institute of Technology (NJIT) and completed a M.S. in
Electrical Engineering at the University of Texas at Austin with a focus on
machine learning applied to wireless communications.
\end{IEEEbiographynophoto}

\begin{IEEEbiographynophoto}{Cheng Lee}
is a PhD student at the University of Texas at Austin advised by Dr.
Joydeep Ghosh. His research work focuses on applications of data
mining and machine learning algorithms in biology and medicine. He
received his B.S. degrees in computer science and electrical
engineering and his M.S. in electrical engineering from the University
of Texas at Dallas. He was also a computational biologist in McDermott
Center for Human Growth and Development at the University of Texas
Southwestern Medical Center.
\end{IEEEbiographynophoto}

\begin{IEEEbiographynophoto}{Joydeep Ghosh}
received the B.Tech. degree from the Indian Institute of Technology Kanpur in
1983 and the Ph.D. degree from the University of Southern California in 1988.
He is currently the Schlumberger Centennial Chair Professor with the Department
of Electrical and Computer Engineering, The University of Texas at Austin,
Austin, where he has been with the faculty since 1988. He has published more
than 250 refereed papers and 50 book chapters, coedited 20 books, and received
14 “best paper” awards.
\end{IEEEbiographynophoto}

\vfill

\end{document}